\documentclass{article}

\PassOptionsToPackage{numbers, compress}{natbib}

\usepackage[preprint]{neurips_2024}
\usepackage{sidecap}
\usepackage[utf8]{inputenc} 
\usepackage[T1]{fontenc}    
\usepackage{hyperref}       
\usepackage{url}            
\usepackage{booktabs}       
\usepackage{amsfonts}       
\usepackage{nicefrac}       
\usepackage{microtype}      
\usepackage{xcolor}         
\usepackage{graphicx}
\usepackage{subcaption}
\usepackage{caption}
\usepackage{float}
\usepackage{amsmath}
\usepackage{mathtools}
\usepackage{tikz}
\usepackage{bm}
\usepackage{pgfplots}      
\usepackage{multirow}
\usepackage{diagbox} 
\usepackage{eucal}
\pgfplotsset{compat=newest} 
\usetikzlibrary{datavisualization}
\usepackage[linesnumbered,ruled,vlined]{algorithm2e}
\usepackage[capitalize]{cleveref}
\crefname{section}{Sec.}{Secs.}
\Crefname{section}{Section}{Sections}
\Crefname{table}{Table}{Tables}
\crefname{table}{Tab.}{Tabs.}
\usepackage{bbm}
\usepackage{amssymb}
\usepackage{bbding}
\usepackage{color}
\usepackage{enumitem}

\usepackage{authblk}
\usepackage{mathrsfs}
\newtheorem{theorem}{Theorem}[section]

\newtheorem{definition}[theorem]{Definition}
\newtheorem{assumption}[theorem]{Assumption}

\title{Improving Generalization of Deep Neural Networks\\ by Optimum Shifting}

\author{
  Yuyan Zhou$^{1}$, Ye Li$^{1}$\thanks{Corresponding author}\;, Lei Feng$^{2}$, Sheng-Jun Huang$^{1}$\\
  $^1$MIIT Key Laboratory of Pattern Analysis and Machine Intelligence, College of Computer Science and Technology, Nanjing University of Aeronautics and Astronautics, Nanjing 211106, China.\\
  $^2$Information Systems Technology and Design Pillar, Singapore University of Technology and Design \\
  \texttt{\{yuyanzhou, yeli20, huangsj\}@nuaa.edu.cn}, \texttt{lfengqaq@gmail.com} \\
}

\begin{document}

\maketitle

\begin{abstract}
Recent studies showed that the generalization of neural networks is correlated with the sharpness of the loss landscape, and flat minima suggests a better generalization ability than sharp minima.
In this paper, we propose a novel method called \emph{optimum shifting}, which changes the parameters of a neural network from a sharp minimum to a flatter one while maintaining the same training loss value.
Our method is based on the observation that when the input and output of a neural network are fixed, the matrix multiplications within the network can be treated as systems of under-determined linear equations, enabling adjustment of parameters in the solution space, which can be simply accomplished by solving a constrained optimization problem.
Furthermore, we introduce a practical stochastic optimum shifting technique utilizing the Neural Collapse theory to reduce computational costs and provide more degrees of freedom for optimum shifting. 
Extensive experiments (including classification and detection) with various deep neural network architectures on benchmark datasets demonstrate the effectiveness of our method. 
\end{abstract}

\section{Introduction}
\label{intro}
Deep neural networks (DNNs) are powerful and have shown remarkable results in various fields, including computer vision~\cite{gans, dpm, vae} and natural language processing~\cite{gpt4, transformer}. 
Contemporary deep learning approaches formulate the learning problem as an optimization problem and utilize stochastic gradient descent and its variants to minimize the loss function.
%
%
%
Nowadays, DNNs are generally overparameterized and capable of providing a larger hypothesis space with normally better solutions having smaller training errors. 
However, this expansive hypothesis space is concurrently populated with different minima, each characterized by distinct generalization abilities.
Recent studies have shown that the generalization of DNNs is correlated with the sharpness of
the loss landscape and flat minima suggest a better generalization ability than
sharp minima~\cite{minima, neyshabur2017implicit, hochreiter1994simplifying, keskar2017large,chaudhari2019entropy,gatmiry2023inductive}.
%
%
%
Existing works on sharpness minimization mainly focus on penalty-based methods ~\cite{SAM, asam, zhaopenalizing}. However, in this work, we propose a penalty-free sharpness minimization technique aimed at addressing a fundamental question below:
\begin{center}
	\textit{Can we change the solution of a DNN from one point to a flatter one while maintaining \\the same training loss value?}
\end{center}

In this paper, we propose a method called \emph{optimum shifting} (OS) to attain this objective.
%
%
This approach is based on the matrix multiplication within the neural network:
\begin{equation}
    \boldsymbol{A} \bm{V} = \boldsymbol{Z},
    \label{1}
\end{equation}
where $\boldsymbol{A}\in \mathbb{R}^{batch\times m}$ represents the input matrix, and $\bm{V}\in \mathbb{R}^{m\times n}$ denotes the parameters in the neural network's linear layer. 
Consider it to be a set of linear equations, and the parameter $\bm{V}$ can be modified in the solution space.
Assume that the rows in $\boldsymbol{A}$ are independent without loss of generality,
if the equation $\boldsymbol{A}\bm{V} = \boldsymbol{Z}$ is under-determined, it will have infinite solutions for $\bm{V}$.
This property allows us to change the neural network parameters from the current point $\bm{V}$ in the solution space to another point $\bm{V}^*$ by minimizing the sharpness (i.e., the trace of its Hessian~\cite{gatmiry2023inductive,pyhessain,blanc2020implicit}), which can be calculated by solving a simple constrained optimization problem.
%

The main challenge for the method above is that it requires keeping training loss unchanged across all training samples.
Using the entire training set for OS demands a significantly large memory capacity and huge computational costs. 
Moreover, OS in the whole dataset results in a less under-determined input matrix, which limits the degrees of freedom for optimum shifting.
To overcome these challenges, we take inspiration from stochastic gradient descent, which uses a small batch to conduct gradient descent, thereby reducing computational costs. 
We propose stochastic optimum shifting, which performs OS on a small batch of data. 
In this way, the generalization ability of the neural network can be improved and the computational costs can be decreased. 
According to the theory of the Neural Collapse~\cite{nc2, tirer2022extended, zhou2022optimization}, \textit{if the loss is unchanged in a small batch of training data (typically, ``$batch\geq n$" where $n$ is the class number), it is expected to be unchanged across all training data}. 
Therefore, the empirical loss is expected to remain unchanged with the stochastic optimum shifting algorithm. 
To summarize, our contributions include:
\begin{itemize}[leftmargin=0.5cm,topsep=-1pt]
    \item We propose optimum shifting (OS), which enables us to change the parameters of neural networks from a sharp minimum to a flatter one while maintaining the same training loss. We prove that the generalization can be improved from the Hessian trace perspective.
    \item 
    We conduct extensive experiments with various deep neural network architectures on benchmark datasets.
    Experimental results show that by using stochastic OS both during training and after training, the deep models can achieve better generalization performance.
    \item Our proposed OS is versatile and can be easily integrated into traditional regularization techniques, such as weight decay and recently proposed methods to find flatter minima (e.g., the sharpness-aware minimization method~\cite{SAM}). 
\end{itemize}

\section{Related work} 
\label{rw}
\textbf{Sharpness and Generalization.}
Research on the correlation between the generalization performance and the sharpness of the loss landscape can be traced back to Hochreiter et al.~\cite{hochreiter1994simplifying}.
Jiang et al.~\cite{jiang2019fantastic} and Yao et al.\cite{pyhessain} performed a large-scale empirical study on various generalization measures and showed that the sharpness of the loss landscape is the most suitable measure correlated with the generalization performance.
Keskar et al.~\cite{keskar2017large} observed that when increasing the batch size of SGD to train a model, the test error and the sharpness would both increase. 
Gatmiry et al.~\cite{gatmiry2023inductive} showed that with the standard restricted isometry property on the measurement, minimizing the trace of Hessian can lead to better generalization.
It is also noteworthy that Dinh et al.~\cite{dinh} argued that for networks with scaling invariance, there always exist models with good generalization but with large sharpness, but this does not contradict with the sharpness minimization algorithms~\cite{gatmiry2023inductive}. 

\textbf{Sharpness Minimization Algorithm.}
Although the above rescaling trick can maximize sharpness and maintain both the test and training loss, it does not contradict with current sharpness minimization method, which only asserts the solution with a minimal trace of Hessian generalizes well, but not vice versa.
Therefore, recent studies have proposed several penalty-based sharpness regularization methods to improve the generalization.
SAM~\cite{SAM} was proposed to penalize the sharpness of the loss landscape to improve the generalization.
The full-batch SAM aims to minimize worst-direction sharpness (Hessian spectrum) and 1-SAM aims to minimize the average-direction sharpness (Hessian trace)~\cite{wen2023sharpness}. 
Furthermore, Kwon et al.~\cite{asam} proposed ASAM, where optimization could keep invariant to a specific weight-rescaling operation. 
In addition, Zhao et al.\cite{zhaopenalizing} proposes to improve the generalization by penalizing the gradient norm during optimization. 
It is worth noting that the methods above are all penalty-based methods, i.e. adding a penalty term to the loss function, to encourage the flatness. 
Compared with them, our method is a constraint and objective separation method, which separates the flatness and loss value as two isolated objectives to optimize.
As a result, our OS can be easily integrated into penalty-based methods (e.g. weight decay) to achieve better performance.

\textbf{Neural Collapse.}
Recent seminal works empirically demonstrated that last-layer features and classifiers of a trained DNN exhibit an intriguing Neural Collapse ($\CMcal{N C}$) phenomenon \cite{nc2, tirer2022extended, zhou2022optimization}.
Specifically, it has been shown that the learned features (the output of the penultimate layer) of within-class samples converge to their class means.
Moreover, $\CMcal{N C}$ seems to take place regardless of the choice of loss functions.
We utilize this phenomenon and propose stochastic optimum shifting, which can reduce the computational costs and provide more degrees of freedom for optimum shifting. 
For example, when we apply optimum shifting to ResNet/DenseNet for CIFAR-100 classification, by ensuring that the loss remains unchanged across 100 samples, the loss remains probabilistically unchanged on the whole 50,000 training data points.

\section{Optimum Shifting}
\noindent\textbf{Notations.}
we denote $\mathcal{S} \triangleq \{(\boldsymbol{x}_i, \boldsymbol{y}_i)\}_{i=1}^n$  as the training set containing $n$ training samples, $L$ as the loss function, and $f_k(\boldsymbol{x}_i)$ as the $k$ layer neural network approximation.
The vectorized output of $i$-th layer is represented using $\text{vec}(\boldsymbol{x}_{l, i})\in \mathbb{R}^{ m_l}$. We vectorize parameters in each layer and stack it as $\mathbf{W} = [\text{vec}(\mathbf{v}), \text{vec}(\mathbf{F}_l), \cdots, \text{vec}(\mathbf{F}_1))]$

\vspace{-.2cm}
\subsection{Motivation}
\label{main results}
\vspace{-.1cm}
Before proceeding, we first define the flatness of neural networks.
There are many measures to define the flatness. But currently, the trace of Hessian has been theoretically proved with the generalization bound~\cite{gatmiry2023inductive}. 
In this paper, we define flatness by the Hessian trace.
\begin{definition}
    The flatness $\mathrm{Flat}(L)$ is defined as the trace of the Hessian matrix $\mathbf{H}_{L}$ of the loss function $L$ with respect to the network parameters $\mathbf{W}$:
    \vspace{-.1cm} 
\begin{align}
    \mathrm{Flat}(L) \triangleq \mathrm{tr}(\mathbf{H}_{L})=  \mathrm{tr}(\nabla_{\boldsymbol{\mathbf{W}}}^2L(\boldsymbol{f}(\boldsymbol{x}_i), \boldsymbol{y}_i)).
\end{align}
\end{definition}
\vspace{-.2cm} 
Thus optimum shifting (OS) can be represented as:
\begin{align}
 \label{eq4}
\vspace{-1.5cm} 
	\text{minimize} \quad \text{tr}(\nabla_{\boldsymbol{\bm{V}}}^2L(\boldsymbol{f}(\boldsymbol{x}_i), \boldsymbol{y}_i)), \quad
	\text{subject to} \quad \boldsymbol{A}\boldsymbol{V} =\boldsymbol{Z}.
 \vspace{-1.5cm} 
\end{align}
However, since~\cref{eq4} dose not closed-form solution, we find its upper and lower bound to optimize. 
Next, we show the main theorem of our paper. It shows that for different neural networks such as CNN, ResNet, DenseNet and MLP, the lower bound and the upper bound of the Hessian trace are linear with the Frobenius norm of the weight in the final linear layer. 
Therefore, when the output of training data is fixed, if we minimize the Frobenius norm of the weight in the final linear layer, both the upper bound and the lower bound of the Hessian trace will also be minimized, thereby suggesting a better generalization ability , which is shown in the following Theorem.
\begin{theorem}\label{thm1}
	For an $l$-layer MLP or convolutional (CNN, ResNet and Densenet) neural network, given the loss function $L$, the trace of Hessian can be bounded by:
	\begin{gather}
		C_0 + C_1\Vert\bm{V}\Vert^2\leq \mathrm{tr}(\mathbf{H}_{L}) \leq C_0 + C_2\Vert \bm{V}\Vert^2,
	\end{gather}
        where $\bm{V}$ is the weight matrix of the last linear layer and $C_0, C_1, C_2$ are constants and independent of the last hidden layer's weight $\bm{V}$. Therefore, if $\Vert \bm{V}\Vert^2$ is minimized, both the upper bound and the lower bound of the Hessian trace will also be minimized.
\end{theorem}
\vspace{-.2cm}
The proof of Theorem \ref{thm1} is included in \cref{appendix a}. 

\subsection{Our Proposed Optimum Shifting Method}
\label{Method}
Motivated by the \cref{thm1} above, we aim to replace the current weight using the least $F$-norm solution in the linear space. 
A linear layer $\mathbb{R}^{m}\rightarrow \mathbb{R}^n$ with the activation function $\sigma$ can be represented as follows:
\begin{equation}
	\phi^{\mathrm{fc}}(\bm{V}) = \sigma(\boldsymbol{A}\bm{V} + \boldsymbol{C}), 
\end{equation}
where $\boldsymbol{A}\in \mathbb{R}^{batch\times m}$, $\bm{V}\in \mathbb{R}^{m\times n}$, $\boldsymbol{C}\in \mathbb{R}^{batch\times n}$ and $batch$ is the batch size. We denote the result of $\boldsymbol{A}\bm{V}$ as 
\begin{equation}
	\boldsymbol{A}\bm{V} \coloneqq \boldsymbol{Z}.
\end{equation} 
The training loss value of a neural network will not change no matter how the parameter $\bm{V}$ is adjusted in the solution space if the input matrix $\boldsymbol{A}$ and output matrix $\boldsymbol{Z}$ are both fixed. 
Specifically, the equation $\boldsymbol{A}\bm{V} = \boldsymbol{Z}$ defines a system of linear equations. 
If this system is under-determined, then $\bm{V}$ has an infinite number of solutions.
Any option in the solution space is available as a replacement for the current $\bm{V}$.
As stated in Section.\ref{main results}, both the upper bound and the lower bound of the Hessian trace are linear with $\Vert \bm{V}\Vert^2$. If  $\Vert \bm{V}\Vert^2$ is minimized, both the upper bound and the lower bound of the Hessian trace will also be minimized.
As a result, we prefer replacing the current solution with the one that has the least Frobenius norm in the solution space. 
This can be obtained by solving a least-square problem as follows:
\begin{gather}
	\text{minimize} \quad \Vert \bm{V}\Vert^2, \quad \\
	\text{subject to} \quad \bm{A}\bm{V} =\boldsymbol{Z}.
\end{gather}
Thus, we aim to find the point with the smallest Frobenius norm in the solution space to replace the current $\bm{V}$.
To solve the least-square problem, we make two assumptions for the input matrix $\bm{A}$.
\begin{assumption}[Low Rank Assumption]
The rank of the input matrix $\bm{A}$ is smaller than its number of columns. i.e., $\text{rank}(\bm{A}) < m$.
\label{a1}
\end{assumption}
\begin{assumption}
[Row Independence Assumption] The rows of the input matrix $\boldsymbol{A}$ are linearly independent. i.e., $\forall \bm{a}\in \mathbb{R}^{batch}$, $\bm{a}^\top\bm{A}=\bm{0}^\top$ if and only if $\bm{a} = \bm{0}$.
\label{a2}
\end{assumption}
\begin{figure*}[!t]
    \centering
    \begin{minipage}{.52\textwidth}
        \centering
        \begin{algorithm}[H]\small
         \KwIn{Training set $\mathcal{S} \triangleq \cup_{i=1}^n\{(\boldsymbol{x}_i, \boldsymbol{y}_i)\}$, batch size $b_1, b_2$ for SGD and SOS, step size $\gamma>0$.}
         \For{number of training epochs}{
          Sample batch $\mathcal{B} = \{(\boldsymbol{x}_1, \boldsymbol{y}_1), ... (\boldsymbol{x}_{b_2}, \boldsymbol{y}_{b_2})\}$\;
          Compute the input and output matrix \;
          $\boldsymbol{A} = \left[ \boldsymbol{x}_{k, 1}, \boldsymbol{x}_{k, 2}, \cdots , \boldsymbol{x}_{k, b_2} \right] \;
          \boldsymbol{Z} = \left[ \bm{V}^T\boldsymbol{x}_{k, 1},  \bm{V}^T\boldsymbol{x}_{k, 2}, \cdots ,  \bm{V}^T\boldsymbol{x}_{k, b_2} \right]$ \;
          \For{each columns $\bm{V}_i$ in the final linear layer}
           {
    			Conduct Gaussian elimination to make $\boldsymbol{A}$ row-independent: $[\boldsymbol{A}_*, \boldsymbol{Z}_{i*}] = \textit{Gaussian eliminate}([\boldsymbol{A}, \boldsymbol{Z}_i])$ \;
    			Update the parameters: $\bm{V}_i^* = \boldsymbol{A}_*(\boldsymbol{A}_*(\boldsymbol{A}_*)^T)^{-1}\boldsymbol{Z}_{i*}$\;
    	}
          \For{$t=0,1,\cdots,s$}
		  {
			Update all parameters using SGD\;
                $\mathbf{W}_{t}$ = $\mathbf{W}_{t-1} - \gamma\frac{1}{b_1}\sum_{i=1}^{b_1} \nabla_{\mathbf{W}_{t-1}} L$\;
		  }
         }
         \caption{SOS algorithm during training}
         \label{sos}
        \end{algorithm}
    \end{minipage}%
    \begin{minipage}{0.48\textwidth}
        \centering
        \vspace{1cm}
  \includegraphics[width=1\textwidth]{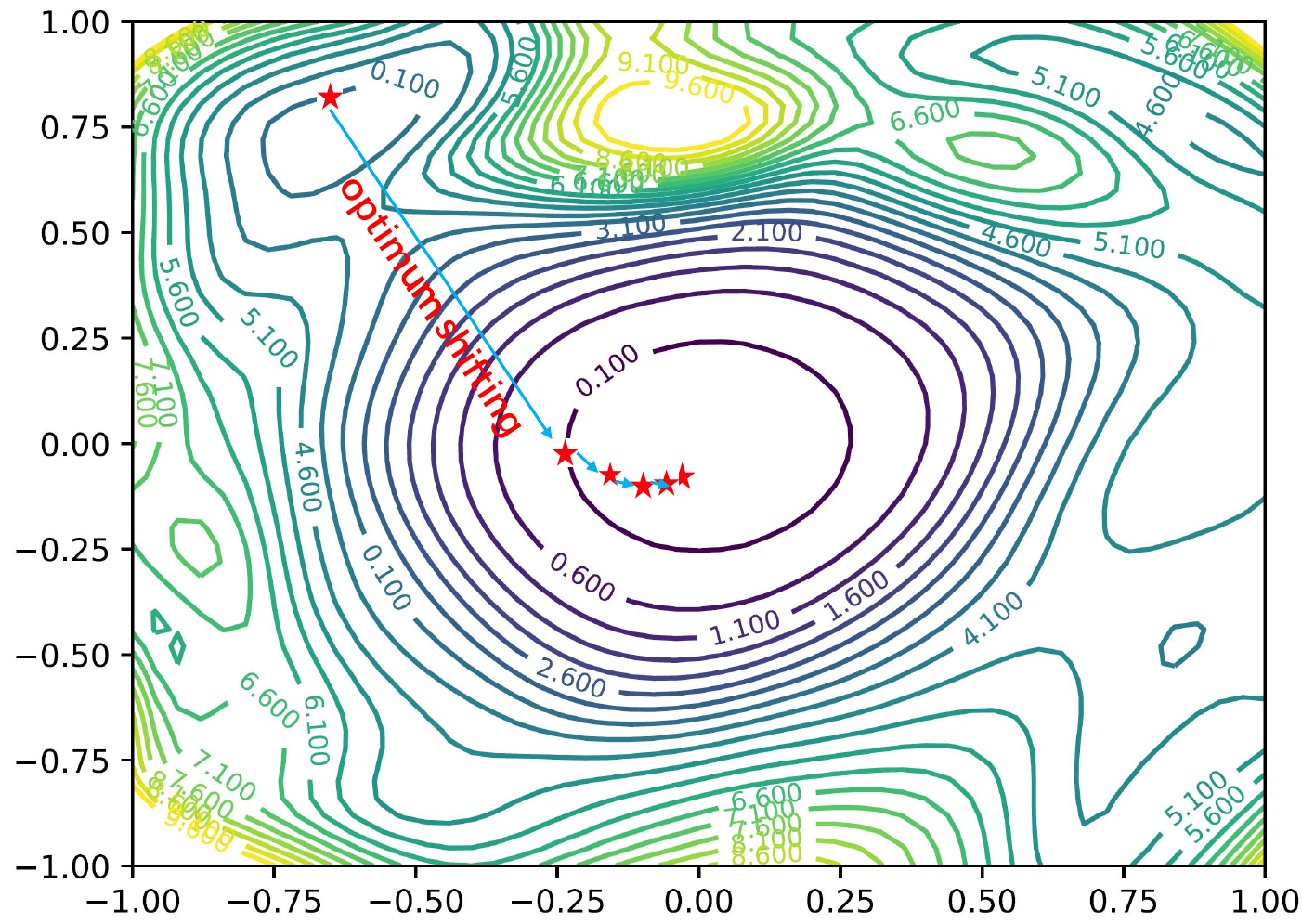}
  \captionof{figure}{Schematic of the OS algorithm.\protect\footnotemark}
  \label{fig:sam-step-schematic}
    \end{minipage}
\end{figure*}
\footnotetext{Figure~\ref{fig:sam-step-schematic} is generated following Li et al.~\cite{li2018visualizing}, which shows the loss landscape for all parameters using 2D plots.}
The first assumption ensures that the linear system is under-determined, and has many solutions in the solution space. The second assumption makes $\bm{A}\bm{A}^\top$ in the closed form solution in \cref{v*} is invertible. In the next section, we will show how to satisfy the two assumptions.

Because $\bm{V}\in \mathbb{R}^{m\times n}$ is a matrix, we need to decompose it into $n$ independent least-square problems.  
We denote the $i$-th column of $\bm{V}$ as $\bm{V}_i$, and the Lagrangian for the least-square problem can be expressed as follows:
\begin{gather}
L_1(\bm{V}_1,\cdots,\bm{V}_n,\boldsymbol{\lambda}_1,\cdots,\boldsymbol{\lambda}_n)
 = \sum\nolimits_{i=1}^n (\bm{V}_i^\top\bm{V}_i + \boldsymbol{\lambda}_i^\top(\boldsymbol{A}\bm{V}_i - \boldsymbol{Z}_i)).
\end{gather}
Since $L_1$ is a convex quadratic function of each variable $(\bm{V}_i,\boldsymbol{\lambda}_i)$, we can find the minimum $(\bm{V}_i^*,\boldsymbol{\lambda}_i^*)$ from the optimality condition:
\begin{align}
	\nabla_{\bm{V}_i} L_1 &= 2\bm{V}_i + \boldsymbol{A}^\top\boldsymbol{\lambda}_i = \boldsymbol{0},\\
\nabla_{\boldsymbol{\lambda}_i}L_1 &= \boldsymbol{A}\bm{V}_i - \boldsymbol{Z}_i = \boldsymbol{0},
\end{align}
which yeilds a closed-form solution $\bm{V}_i^* = \boldsymbol{A}^\top(\boldsymbol{A}\boldsymbol{A}^\top)^{-1}\boldsymbol{Z}_i$. 
By resolving $n$ independent least-square problems, we can finally get the point with the smallest Frobenius norm as 
\begin{equation}
    \bm{V}^* = [\bm{V}^*_1, \bm{V}^*_2, \cdots, \bm{V}_n^*] = \boldsymbol{A}^\top(\boldsymbol{A}\boldsymbol{A}^\top)^{-1}\boldsymbol{Z}. 
    \label{v*}
\end{equation}
In what follows, we will detail how to satisfy Assumption~\ref{a1} and Assumption~\ref{a2}.

\paragraph{Stochastic Optimum Shifting.}
\label{methodsos}
As mentioned in Assumption~\ref{a1}, OS relies on the low-rank property of the input matrix $\bm{A}$. This property ensures that the linear system is under-determined, allowing for an infinite number of solutions within the solution space for OS to change. It is important to highlight that when using a small batch for OS, i.e., $batch < m$, we have 
\begin{gather}
    \text{rank}(A)\leq batch < m,
\end{gather}
which satisfies the requirement specified in Assumption~\ref{a1}.

To achieve optimum shifting, the training loss is expected to remain unchanged across the entire dataset. 
However, for stochastic optimum shifting, we only maintain it unchanged in a small batch.
Neural Collapse~\cite{nc2, tirer2022extended, zhou2022optimization} helps us to understand this phenomenon.
It reveals that: as training progresses, the within-class variation of the activations becomes negligible as they collapse to their class means. 
For example, when training on the CIFAR-100 dataset, the images will converge to the 100 class means. 
Therefore, if the loss of 100 images remains unchanged after performing optimum shifting to the final fully connective layer, the loss of the entire dataset will also remain nearly unchanged, which can both satisfy Assumption~\ref{a1} and reduce the computationl costs.  

Moreover, the small batch for optimum shifting also offers more degrees of freedom for optimum shifting.
For instance, when the last layer maps a vector with $1024$ dimensions to $100$ dimensions, i.e., the weight matrix $\bm{V}\in \mathbb{R}^{1024\times 100}$. 
When feeding the entire dataset to the neural network, the input matrix $\boldsymbol{A}\in \mathbb{R}^{50000\times 1024}$, which may not be under-determined.
When using stochastic optimum shifting with $batch=300$, the input matrix $\boldsymbol{A}\in \mathbb{R}^{300\times 1024}$, which means that the systems of linear equations are under-determined.
Hence it must have infinite solutions in the solution space. 

\paragraph{Gaussian elimination.}
To ensure the invertibility of $\bm{A}\bm{A}^\top$ in the closed-form solution of \cref{v*}, it is imperative to fulfill the condition stated in Assumption~\ref{a2}.
However, in practice, $\bm{A}$ may be row-dependent and $\bm{A}\bm{A}^\top$ may be singular. After Gaussian elimination, each non-zero row of input matrix $\bm{A}$ will be independent and those zero rows will be discarded. Consequently, the resulting matrix $\bm{A}_*$ in Algorithm \ref{sos} becomes row-independent, ensuring that the matrix $\bm{A}_*\bm{A}_*^\top$ is invertible.
\section{Experiments}
\begin{table*}[t]
    \Huge 
    \caption{Test Accuracy on CIFAR Classification for trained models}
    \vskip 0.15in \hspace{-1.2cm}
\resizebox{1.1\textwidth}{!}{
\setlength{\tabcolsep}{6.0mm}{
\renewcommand{\arraystretch}{1.2}
    \begin{tabular}{lcccc}
    \toprule                     
      & \multicolumn{2}{c}{CIFAR-100} & \multicolumn{2}{c}{CIFAR-10}  \\ \midrule
        VGG-16 & \multicolumn{1}{|c}{Basic} & \multicolumn{1}{c|}{Mixup Augmentation} & \multicolumn{1}{|   c}{Basic} & \multicolumn{1}{c}{Mixup Augmentation} \\ \midrule
        Standard        & $\ 72.12_{\pm 0.52}\ $ & $\ 73.22_{\pm 0.41}\ $ & $\ 92.01_{\pm 0.45}\ $  & $\ 93.21_{\pm 0.32}\ $ \\
        \textbf{+ SOS (ours)}              & $\ \textbf{ 72.23}_{\pm 0.54}\ $ & $\ \textbf{73.38}_{\pm 0.81}\ $ & $\ \textbf{92.28}_{\pm 0.22}\ $  & $\ \textbf{93.35}_{\pm 0.43}\ $  \\ \midrule
        ResNet-18 & \multicolumn{1}{|c}{Basic} & \multicolumn{1}{c|}{Mixup Augmentation} & \multicolumn{1}{|c}{Basic} & \multicolumn{1}{c}{Mixup Augmentation} \\
        \midrule
        Standard        & $\ 78.03_{\pm 0.41}\ $ & $\ 79.23_{\pm 0.19}\ $ & $\ 95.31_{\pm 0.26}\ $  & $\ 95.97_{\pm 0.21}\ $  \\
        \textbf{+ SOS (ours)}            & $\ \textbf{78.25}_{\pm 0.37} \ $ & $\ \textbf{79.88}_{\pm 0.76} \ $ & $\ \textbf{95.47}_{\pm 0.26}\ $  & $\ \textbf{96.04}_{\pm 0.25} \ $  \\
        \midrule
        ResNet-50\ \ \ \  & \multicolumn{1}{|c}{Basic} & \multicolumn{1}{c|}{Mixup Augmentation} & \multicolumn{1}{|c}{Basic} & \multicolumn{1}{c}{Mixup Augmentation} \\
        \midrule
        Standard        &  $\ 79.30_{\pm 0.22}\ $ & $\ 79.53_{\pm 0.21}\  $ & $\ 95.30_{\pm 0.16}\ $  & $\ 96.10_{\pm 0.08}\ $  \\
        \textbf{+ SOS (ours)}              & $\ \textbf{79.57}_{\pm 0.24}\ $ & $\ \textbf{79.72}_{\pm 0.31}\ $ & $\ \textbf{95.47}_{\pm 0.28}\  $  & $\ \textbf{96.22}_{\pm 0.21} \ $  \\
        \midrule
        DenseNet-201  & \multicolumn{1}{|c}{Basic} & \multicolumn{1}{c|}{Mixup Augmentation} & \multicolumn{1}{|c}{Basic} & \multicolumn{1}{c}{Mixup Augmentation} \\
        \midrule
        Standard        & $\ 80.02_{\pm 0.32}\ $ & $\ 81.36_{\pm 0.33}\ $ & $\ 95.31_{\pm 0.22}\ $  & $\ 96.06_{\pm 0.06}\ $  \\
        \textbf{+ SOS (ours)}      & $\ \textbf{80.35}_{\pm 0.26}\ $ & $\ \textbf{81.55}_{\pm 0.28}\ $ & $\ \textbf{95.64}_{\pm 0.36}\ $  & $\ \textbf{96.31}_{\pm 0.34} \ $  \\
        \bottomrule
    \end{tabular}
    \label{trained}
    }
    }
\end{table*}
To validate the effectiveness of our proposed SOS, we investigate the performance of different DNNs (VGG, ResNet, DenseNet, Shufflenet, ResNext, Yolo) on different computer vision tasks, including CIFAR-10/100, ImageNet classification and PASCAL VOC object detection.
In the first subsection, SOS is applied to trained deep models. In the second subsection, it is further applied repeatedly in the training process.
When SOS is applied in the training process, we compare our method with two other training schemes: one is the standard SGD training scheme with weight decay and the other is the SAM~\cite{SAM} training scheme.
As we can see below, SOS improves the generalization ability of trained models, standard SGD scheme with weight decay, and SAM training scheme.
\begin{figure*}[t]
    \centering
    \begin{subfigure}{0.32\textwidth}
    \centering
    \hspace{-.5cm}
        \includegraphics[width=1.07\linewidth]{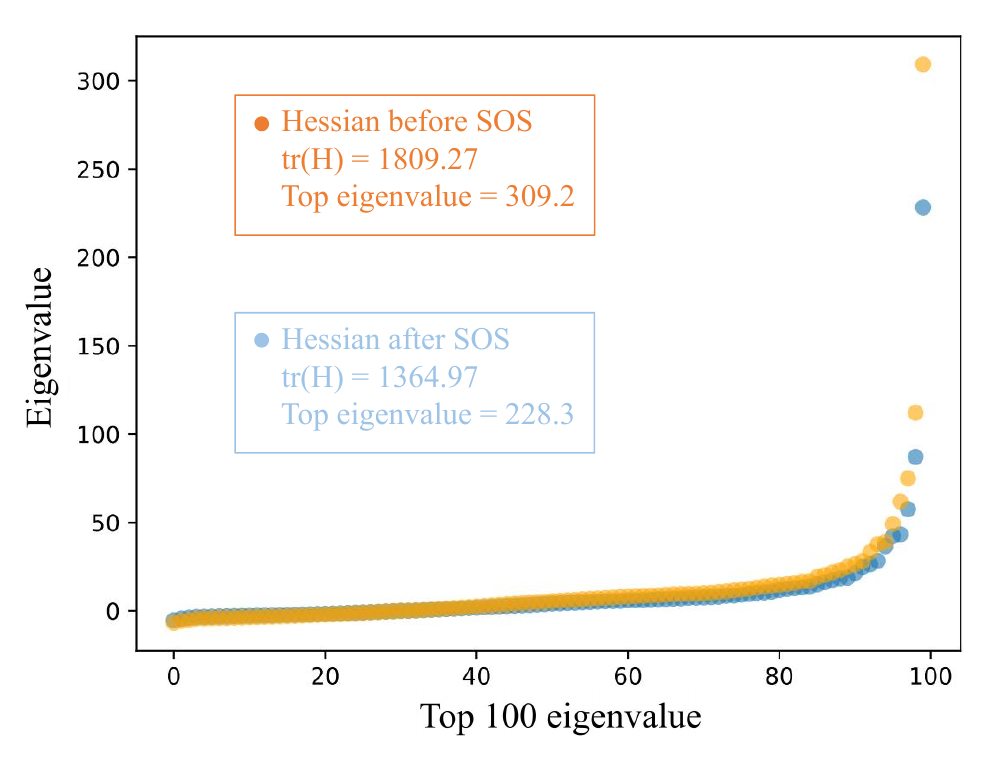}
        \vspace{-.5mm}
        \caption{VGG16 on CIFAR-10}
    \end{subfigure}\hspace{.4mm}
    \begin{subfigure}{0.32\textwidth}
    \centering
        \includegraphics[width=1.03\textwidth]{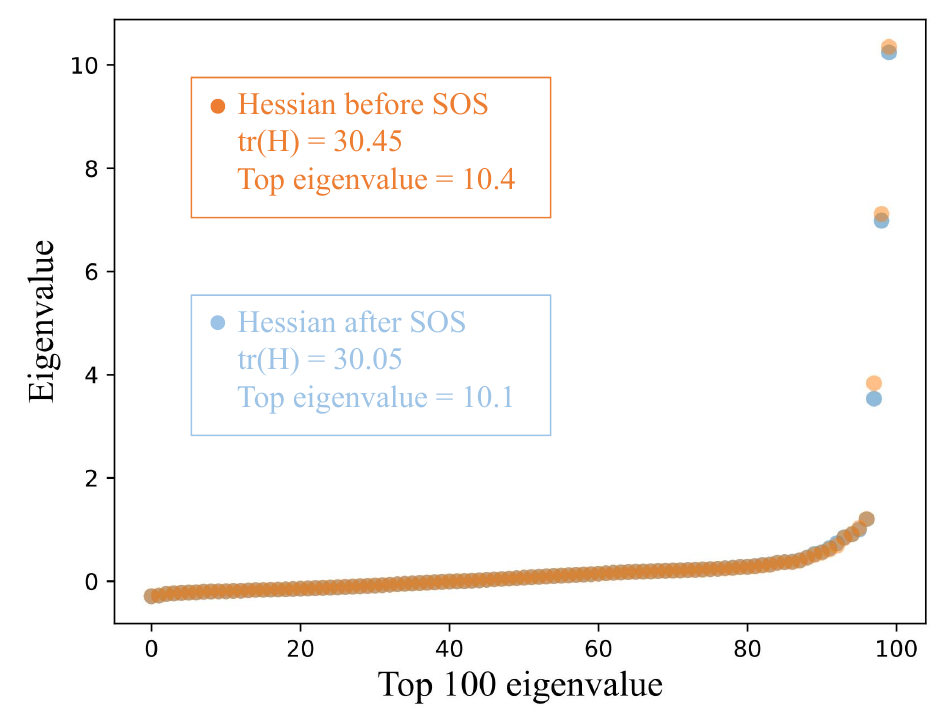}
        \caption{ResNet18 on CIFAR-10}
    \end{subfigure}\hspace{2.5mm}
    \begin{subfigure}{0.32\textwidth}
    \centering
        \includegraphics[width=1.05\textwidth, height=3.5cm]{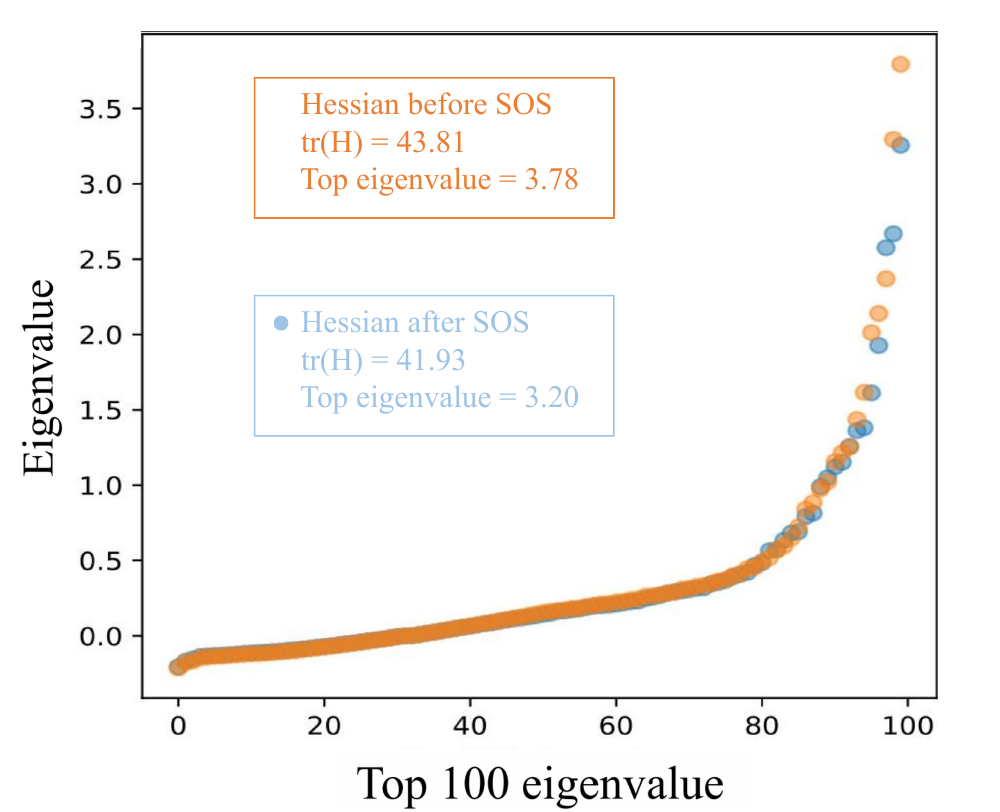}
        \caption{DenseNet121 on CIFAR-10} 
    \end{subfigure}\\
    \hspace{-.1cm}
    \begin{subfigure}{0.32\textwidth}
    \hspace{-.4cm}
    \centering
        \includegraphics[width=1.04\linewidth]{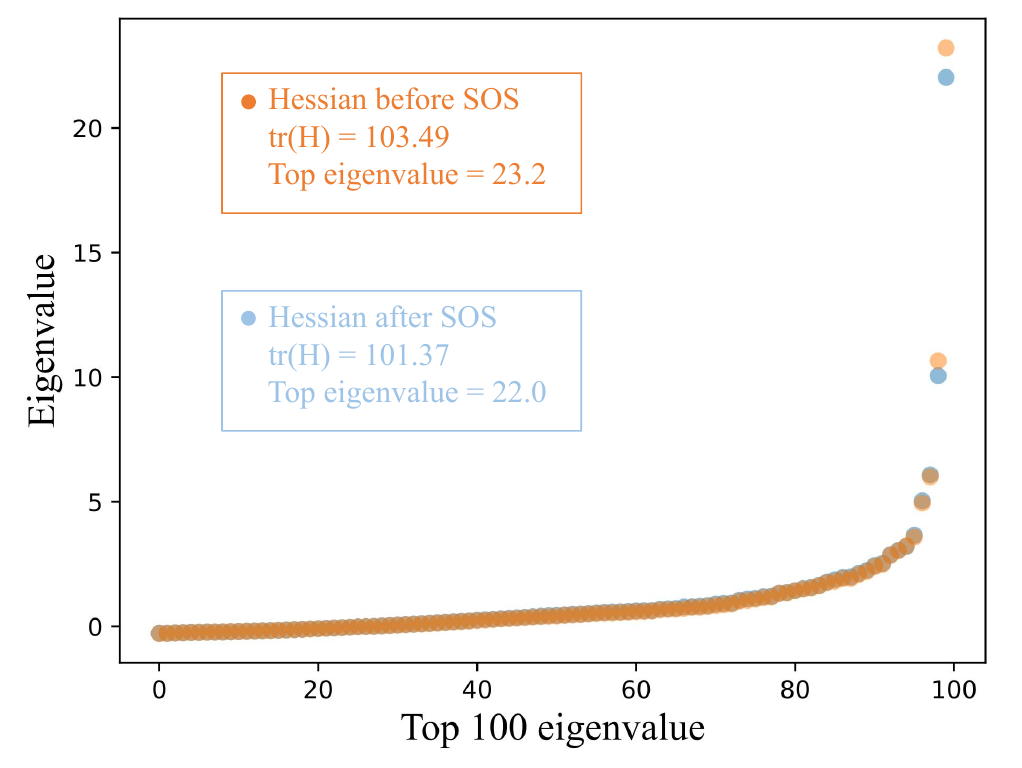}
        \vspace{-.1cm}
        \caption{VGG16 on CIFAR-100}
    \end{subfigure}\hspace{.4mm}
    \begin{subfigure}{0.32\textwidth}
    \centering
        \includegraphics[width=1.04\textwidth]{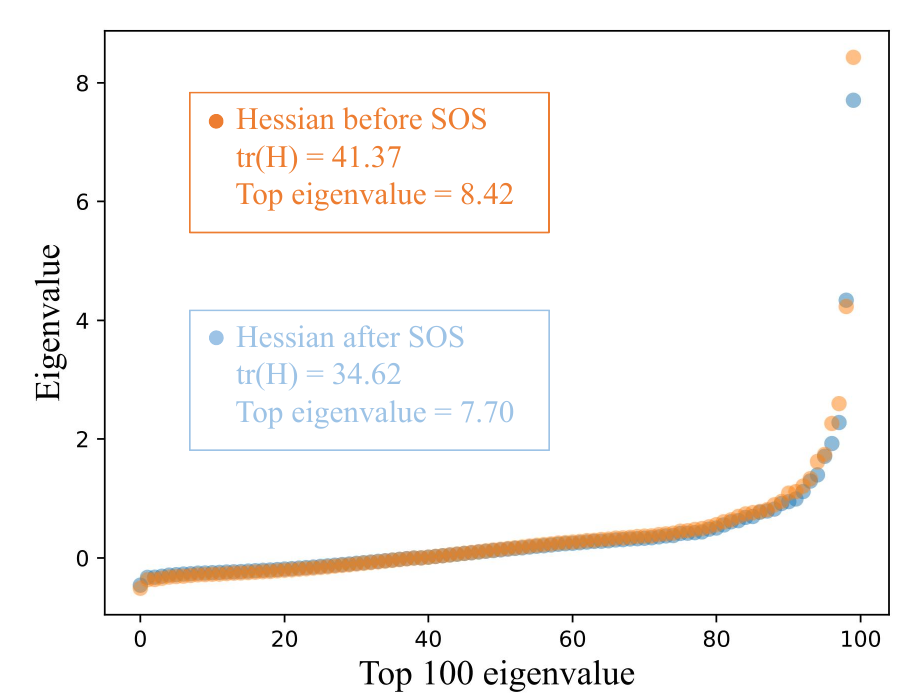}
        \caption{ResNet18 on CIFAR-100}
    \end{subfigure}\hspace{3.5mm}
    \begin{subfigure}{0.32\textwidth}
    \centering
        \includegraphics[width=1.04\textwidth]{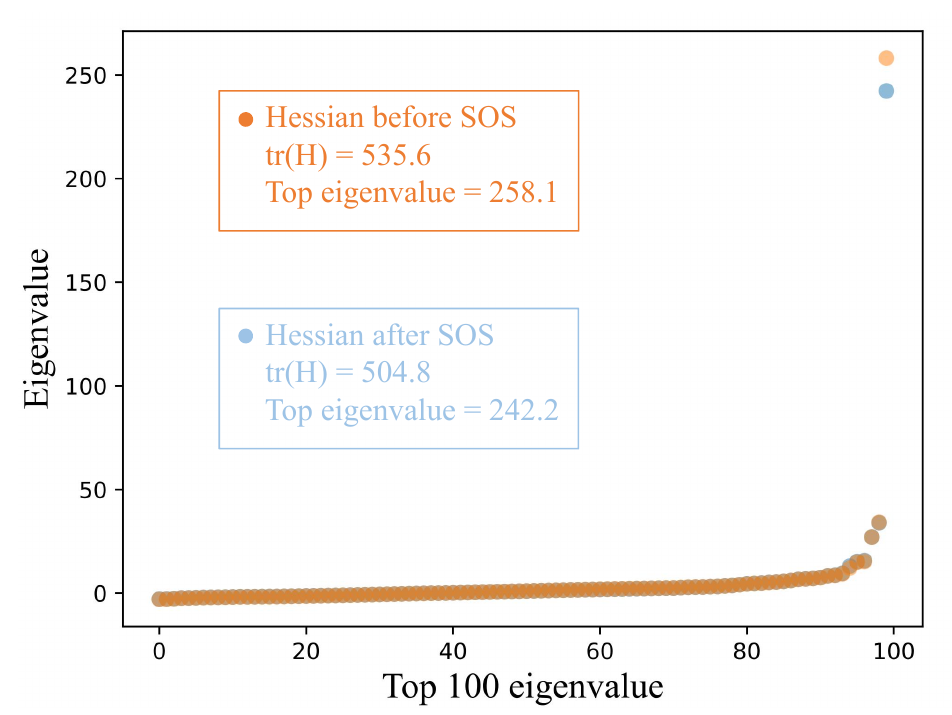}
        \caption{DenseNet121 on CIFAR-100}
    \end{subfigure}
    \caption{Top 100 eigenvalues of the Hessian matrix before and after SOS.}
    \label{Hessian}
\end{figure*}
\subsection{Applying SOS to Trained Models}
\label{ostrained}
\noindent \textbf{Test Accuracy.}
We first evaluate SOS by applying it to trained deep models on CIFAR-10 and CIFAR-100 dataset~\cite{CIFAR10}, which consists of 50k training images and 10k testing images in 10 and 100 classes. 
Different convolutional neural network architectures are tested, including relatively simple architectures, such as VGG~\cite{vgg}, and complex architectures, such as ResNet~\cite{resnet} and DenseNet \cite{dense}.
For the training datasets, we employ data augmentations. One is the basic augmentation (basic normalization and random horizontal flip) and the other is Mixup augmentation~\cite{mixup}. 
%
%
Table~\ref{trained} shows the results. We can see all the test accuracy has been improved slightly by SOS. For example, the test accuracy of VGG-16 on CIFAR-100 has been improved from $72.23_{\pm 0.54}\%$ to $73.38_{\pm 0.81}\%$.

\textbf{Hessian Analysis.}
We visualize the top 100 eigenvalues of the Hessian matrix before and after SOS in ascending order as shown in Figure~\ref{Hessian}. The orange and blue spots represent the eigenvalue before and after SOS. We also calculate the trace and top-1 eigenvalue for analysis. We can see that both the trace (average-direction sharpness) and the top 1 eigenvalue (worst-direction sharpness) have been minimized by SOS, which is consistent with our proposed Theorem~\ref{thm1}.

\noindent\textbf{ImageNet Classification.}
We evaluate our method on the ImageNet classification dataset using three PyTorch official pre-trained deep vision models: ResNet50, Shufflenet\_V2~\cite{dense}, and ResNext50  and apply SOS to test its effectiveness.
Table~\ref{fig:cossim} reports the test accuracy on Imagenet classification with different models.
We can see that SOS also improves both top-1 and top-5 accuracy for different model architectures. For example, for ResNet-50, the top-1 accuracy has been increased from $76.13$\% to $76.61$\% and the top-5 accuracy has been increased from $92.86$\% to $93.14$\%, which further verified our method can also improve the generalization ability in the large scale dataset.

\begin{SCtable}
    \centering
    \vspace{-.5cm}
    \hspace{-1cm}
    \begin{minipage}{0.67\linewidth}
    \hspace{-.2cm}
    \centering
        \begin{tabular}{l|c|cc}
    \toprule
        Network                     &  SOS         &  Top-1 acc  &  Top-5 acc      \\ \midrule
        \multirow{2}{*}{ResNet50}   &  Before SOS &  $76.14$          & $92.86$             \\ 
                                    &  After SOS  &  $\mathbf{76.61}$ & $\mathbf{93.14}$     \\ \midrule
        \multirow{2}{*}{Shufflenet\_V2}    &  Before SOS &  $76.23$          & $93.00$             \\ 
                                           &  After SOS  &  $\mathbf{76.31}$ & $\mathbf{93.04}$     \\ \midrule
        \multirow{2}{*}{ResNext50}  &  Before SOS &  $77.61$          & $93.68$             \\ 
                                    &  After SOS  &  $\mathbf{77.68}$ & $\mathbf{93.72}$     \\ \bottomrule
                                    
    \end{tabular}
  \end{minipage}
    \captionsetup{width=1\textwidth}
    \vspace{1cm}
    \sidecaptionvpos{figure}{t}
    \hspace{-.7cm}
    \caption{\textbf{Applying SOS to pre-trained models on ImageNet dataset.} We also use three PyTorch official pre-trained models: ResNet50, Shufflenet\_V2, and ResNext50 to test the effectiveness of our method. We can see that both the top-1 and top-5 accuracy have been increased by SOS algorithm.}
    \label{fig:cossim}
\end{SCtable}
\begin{figure*}[t]
\vspace{-1cm}
    \begin{subfigure}{0.32\textwidth}
    \hspace{-.7cm}
        \includegraphics[width=1.13\linewidth]{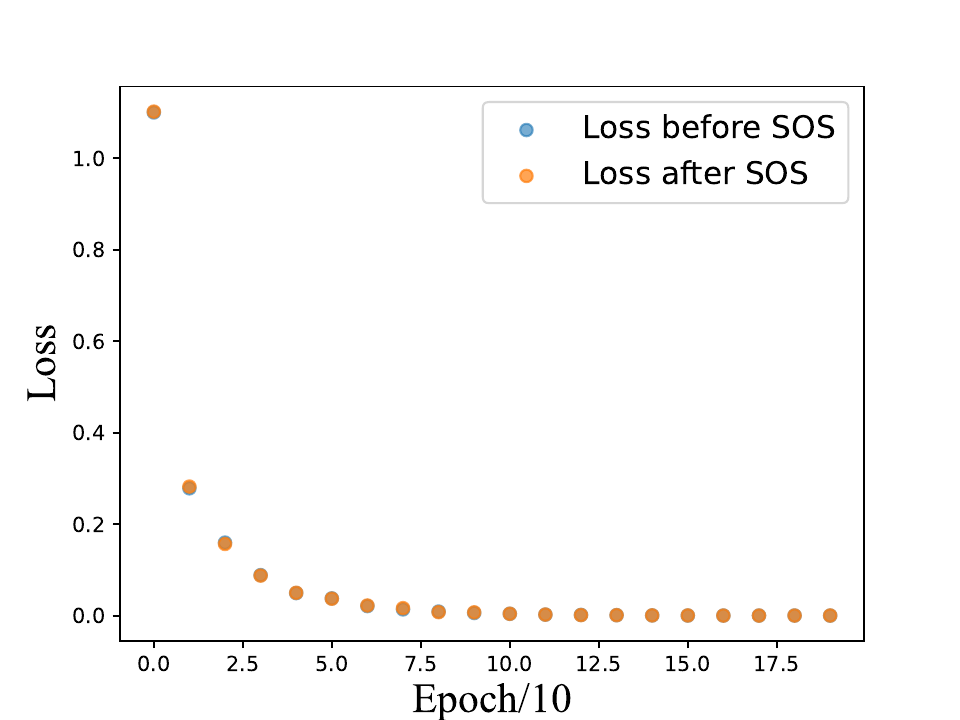}
        \caption{VGG16 on CIFAR-10}
    \end{subfigure}
        \hspace{-.6cm}
    \begin{subfigure}{0.32\textwidth}
        \includegraphics[width=1.13\linewidth]{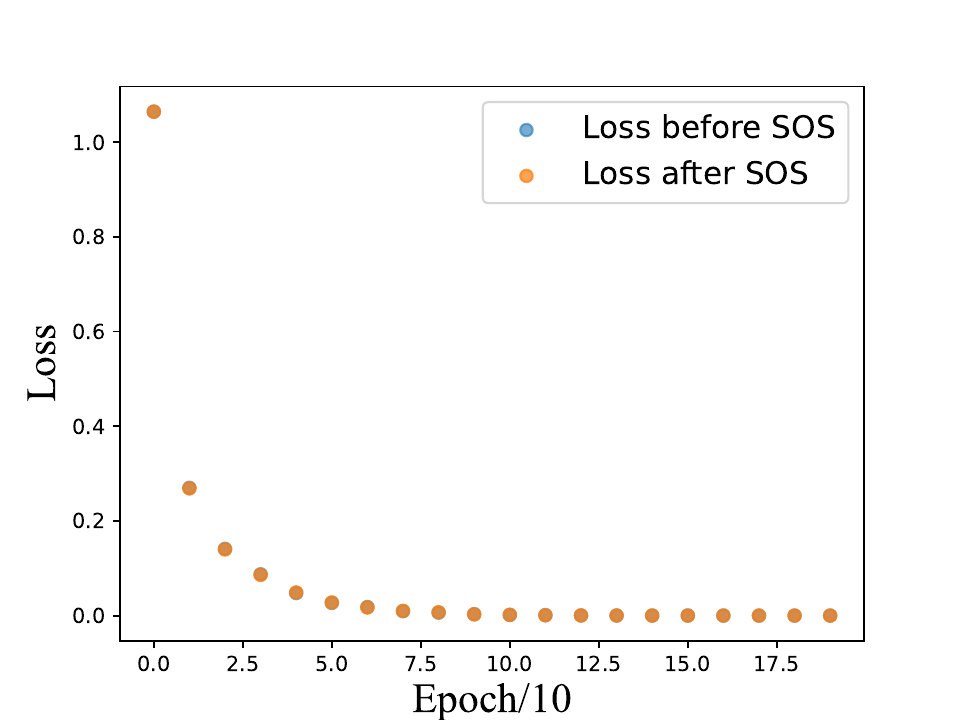}
        \caption{ResNet18 on CIFAR-10}
    \end{subfigure}
    \begin{subfigure}{0.32\textwidth}
        \includegraphics[width=1.13\linewidth]{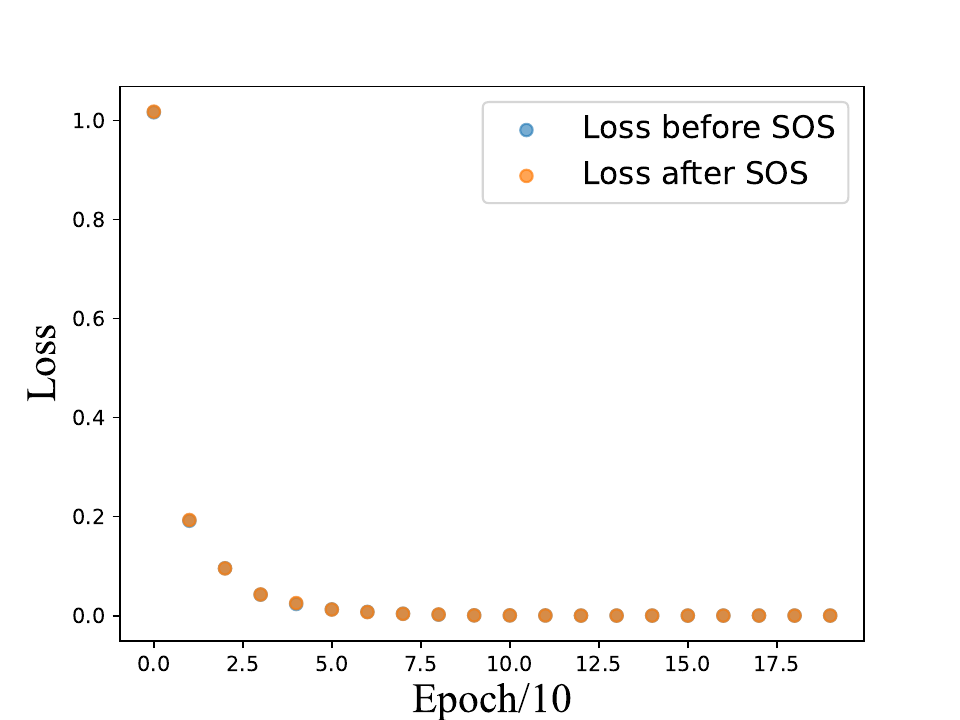}
        \caption{DenseNet121 on CIFAR-10} 
    \end{subfigure}\\
    \begin{subfigure}{0.32\textwidth}
        \hspace{-.7cm}
        \includegraphics[width=1.13\linewidth]{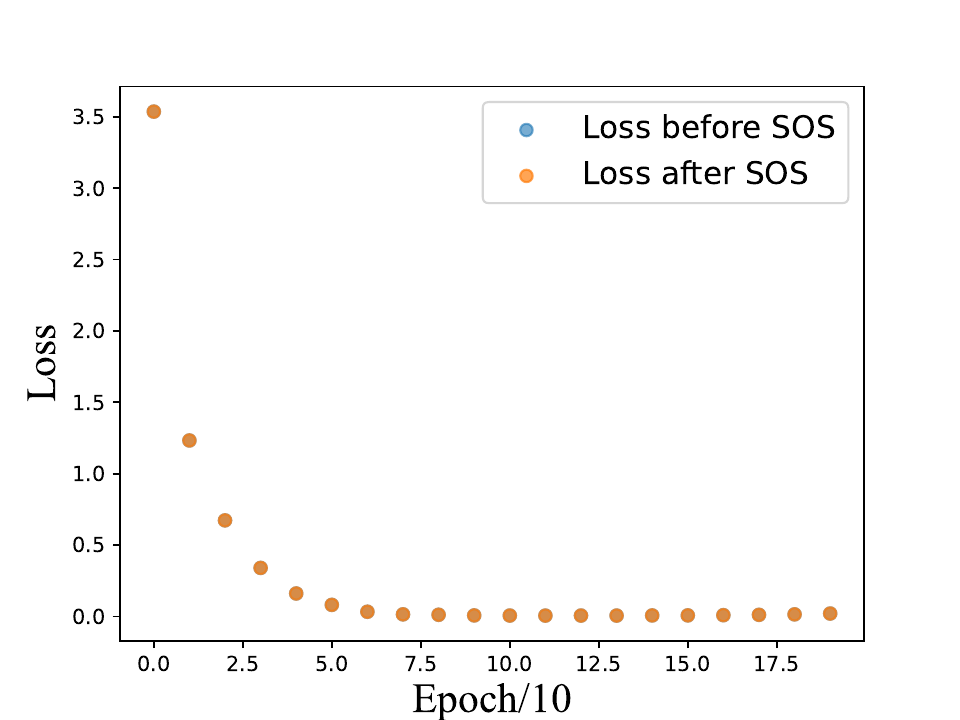}
        \caption{VGG16 on CIFAR-100}
    \end{subfigure}
    \hspace{-.6cm}
    \begin{subfigure}{0.32\textwidth}
        \includegraphics[width=1.13\linewidth]{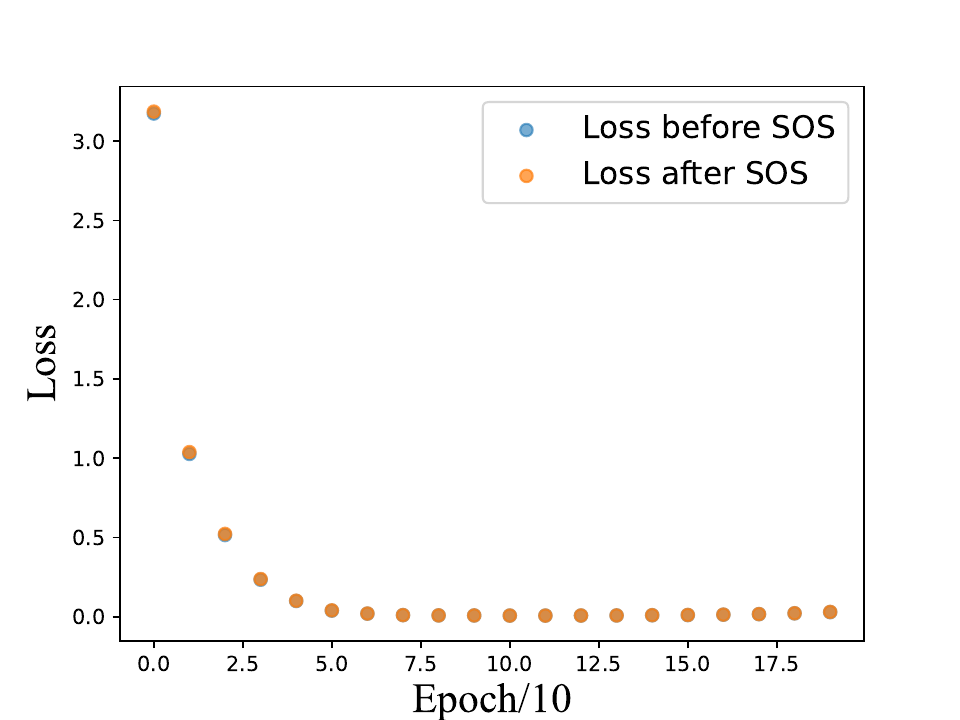}
        \caption{ResNet18 on CIFAR-100}
    \end{subfigure}
    \begin{subfigure}{0.32\textwidth}
        \includegraphics[width=1.13\linewidth]{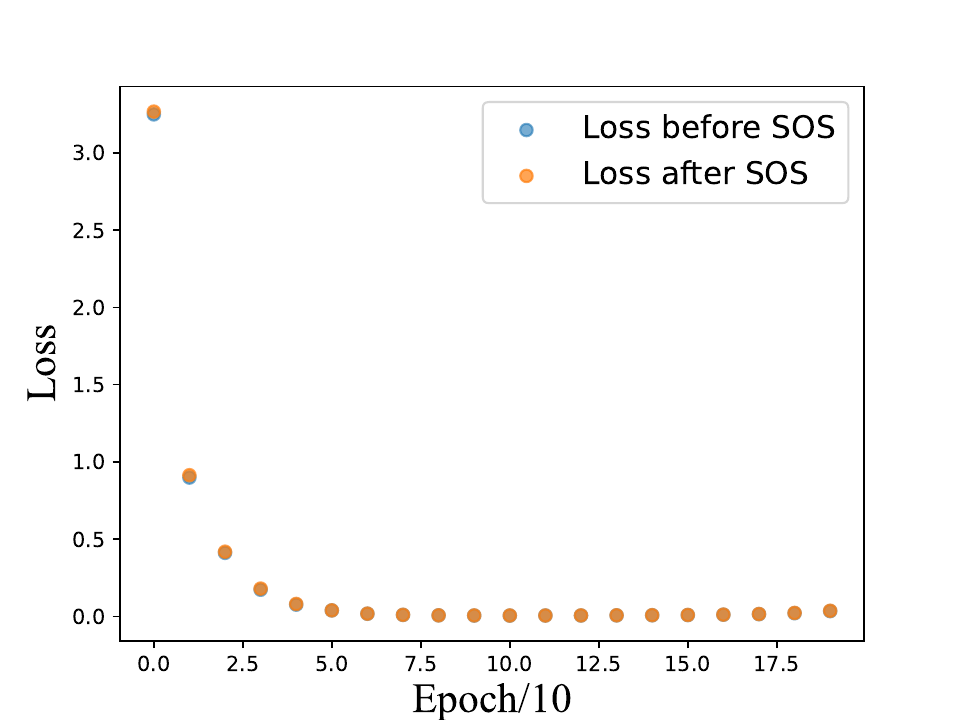}
        \caption{DenseNet121 on CIFAR-100} 
    \end{subfigure}\\
    \caption{Exact loss difference before and after SOS during training.}
    \vspace{-0.1cm}
    \label{Exact Loss}
\end{figure*}

\subsection{Applying SOS in Training Process}
\label{experiment}
\noindent\textbf{Exact Loss difference before/after SOS.} 
The stochastic OS is inspired by the Neural Collapse phenomenon to keep the loss unchanged on a small batch.
Rigorously speaking, we can keep the loss unchanged on CIFAR100 with 100 samples in the latter training stage.
However, in the mid and beginning training phase, the data points will not converge to the class mean. 
So, we used more samples (for example: 300 in the experiments for CIFAR100) and hoped that this would also keep the loss unchanged in the mid and beginning training phases.
Moreover, we have implemented a new experiment about it and show the result below. We first train a ResNet50 on CIFAR100 and save the checkpoint at 10, 20, $\ldots$, 200 epochs. 
Then, we apply SOS to those checkpoints and record the training loss before and after SOS.
We can see in Figure.~\ref{Exact Loss}, at an early stage, the loss after SOS increases very slightly. 
As the training progressed on, the loss remained unchanged in the later epochs.
\begin{table*}[htb]
    \caption{Testing accuracy of different CNN models on CIFAR-10 and CIFAR-100 when implementing the four training schemes.}
    \vskip 0.15in \hspace{-.5cm}
\resizebox{1.05\textwidth}{!}{
\setlength{\tabcolsep}{5.0mm}{
\renewcommand{\arraystretch}{1.2}
    \begin{tabular}{lcccc}
    \toprule
     & \multicolumn{2}{|c|}{CIFAR-100} & \multicolumn{2}{|c}{CIFAR-10} \\
    \midrule
    VGG-16 & \multicolumn{1}{|c}{Basic} & \multicolumn{1}{c|}{Mixup Augmentation} & \multicolumn{1}{|   c}{Basic} & \multicolumn{1}{c}{Mixup Augmentation} \\
    \midrule
    SGD        & $\ 72.12_{\pm 0.52}\ $ & $\ 73.22_{\pm 0.41}\ $ & $\ 92.01_{\pm 0.45}\ $  & $\ 93.21_{\pm 0.32}\ $  \\
    \textbf{+ SOS (ours)}              & $\ \textbf{73.35}_{\pm 0.43}\ $ & $\ \textbf{74.36}_{\pm 0.72}\ $ & $\ \textbf{92.37}_{\pm 0.34}\ $  & $\ \textbf{93.62}_{\pm 0.26}\ $  \\
    \midrule
    SAM             & $\ 74.63_{\pm 0.12}\ $ & $\ 74.96_{\pm 0.56}\ $ & $\ 94.47_{\pm 0.33}\ $  & $\ 94.82_{\pm 0.52}\ $  \\    
    \textbf{+ SOS (ours)}          & $ \ \textbf{74.92}_{\pm 0.03}\ $ & $\ \textbf{75.38}_{\pm 0.29}\ $ & $\ \textbf{94.76}_{\pm 0.47}\ $  & $\ \textbf{95.10}_{\pm 0.31}\ $  \\
    \midrule
    ResNet-18 & \multicolumn{1}{|c}{Basic} & \multicolumn{1}{c|}{Mixup Augmentation} & \multicolumn{1}{|c}{Basic} & \multicolumn{1}{c}{Mixup Augmentation} \\
    \midrule
    SGD        & $\ 78.03_{\pm 0.41}\ $ & $\ 79.23_{\pm 0.19}\ $ & $\ 95.31_{\pm 0.26}\ $  & $\ 95.97_{\pm 0.21}\ $  \\
    \textbf{+ SOS (ours)}            & $\ \textbf{78.43}_{\pm 0.24} \ $ & $\ \textbf{79.97}_{\pm 0.46} \ $ & $\ \textbf{95.54}_{\pm 0.21}\ $  & $\ \textbf{96.17}_{\pm 0.04} \ $  \\
    \midrule
    SAM             & $\ 78.63_{\pm 0.33}\ $ & $\ 80.18_{\pm 0.12}  \ $ & $\ 95.94_{\pm 0.24} \ $  & $\ 96.11_{\pm 0.32}\ $  \\    
    \textbf{+ SOS (ours)} & $\ \textbf{79.66}_{\pm 0.18}\ $ & $\ \textbf{80.43}_{\pm 0.23}\ $ & $\ \textbf{96.18}_{\pm 0.11}\ $  & $\ \textbf{96.48}_{\pm 0.51}\ $  \\
    \midrule
    ResNet-50\ \ \ \  & \multicolumn{1}{|c}{Basic} & \multicolumn{1}{c|}{Mixup Augmentation} & \multicolumn{1}{|c}{Basic} & \multicolumn{1}{c}{Mixup Augmentation} \\
    \midrule
    SGD        &  $\ 79.30_{\pm 0.22}\ $ & $\ 79.53_{\pm 0.21}\  $ & $\ 95.30_{\pm 0.16}\ $  & $\ 96.10_{\pm 0.08}\ $  \\
    \textbf{+ SOS (ours)}              & $\ \textbf{79.63}_{\pm 0.27}\ $ & $\ \textbf{79.90}_{\pm 0.43}\ $ & $\ \textbf{95.53}_{\pm 0.23}\  $  & $\ \textbf{96.33}_{\pm 0.12} \ $  \\
    \midrule
    SAM             & $\ 78.62_{\pm 0.36}\ $ & $\ 81.65_{\pm 0.27}\ $ & $\ 95.82_{\pm 0.18}\ $  & $\ 96.51_{\pm 0.27}\ $  \\    
    \textbf{+ SOS (ours)}   & $\ \textbf{78.97}_{\pm 0.19}\ $ & $\ \textbf{82.41}_{\pm 0.33}\ $ & $\ \textbf{96.31}_{\pm 0.21}\ $  & $\ \textbf{96.88}_{\pm 0.30}\ $  \\
    \midrule
    DenseNet-201  & \multicolumn{1}{|c}{Basic} & \multicolumn{1}{c|}{Mixup Augmentation} & \multicolumn{1}{|c}{Basic} & \multicolumn{1}{c}{Mixup Augmentation} \\
    \midrule
    SGD        & $\ 80.02_{\pm 0.32}\ $ & $\ 81.36_{\pm 0.33}\ $ & $\ 95.31_{\pm 0.22}\ $  & $\ 96.06_{\pm 0.06}\ $  \\
    \textbf{+ SOS (ours)}               & $\ \textbf{80.43}_{\pm 0.13}\ $ & $\  \textbf{81.60}_{\pm 0.14}\ $ & $\ \textbf{95.83}_{\pm 0.12}\ $  & $\ \textbf{96.40}_{\pm 0.21} \ $  \\
    \midrule
    SAM             & $\ 80.14_{\pm 0.56}\ $ & $\ 82.78_{\pm 0.42}\ $ & $\ 96.08_{\pm 0.25} \ $  & $\ 96.82_{\pm 0.14}\ $  \\    
    \textbf{+ SOS (ours)}   & $\ \textbf{81.06}_{\pm 0.23}\ $ & $\ \textbf{83.13}_{\pm 0.13}\ $ & $\ \textbf{96.47}_{\pm 0.32}\ $  & $\ \textbf{97.05}_{\pm 0.42}\ $  \\
    \bottomrule
    \end{tabular}
    }
    }
    \label{accuray_table}
\end{table*}
\begin{figure}[htb]
    \centering
    \begin{subfigure}{0.32\textwidth}
    \centering
    \includegraphics[width=0.95\textwidth]{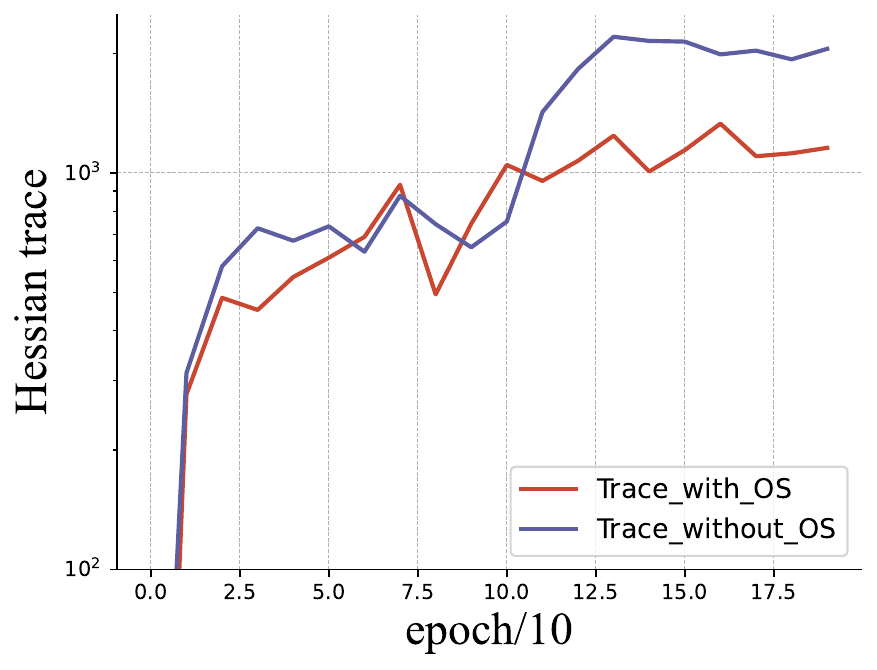}
    \subcaption{VGG}
    \end{subfigure}
    \begin{subfigure}{0.32\textwidth}
    \centering
    \includegraphics[width=0.95\textwidth]{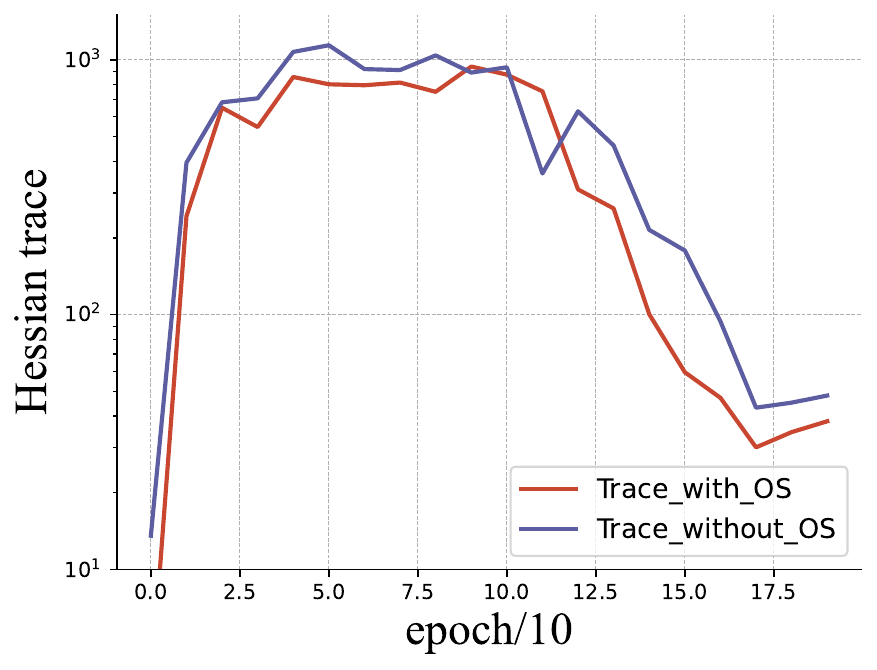}
    \subcaption{ResNet18}
    \end{subfigure}
    \begin{subfigure}{0.32\textwidth}
    \centering
    \includegraphics[width=0.95\textwidth]{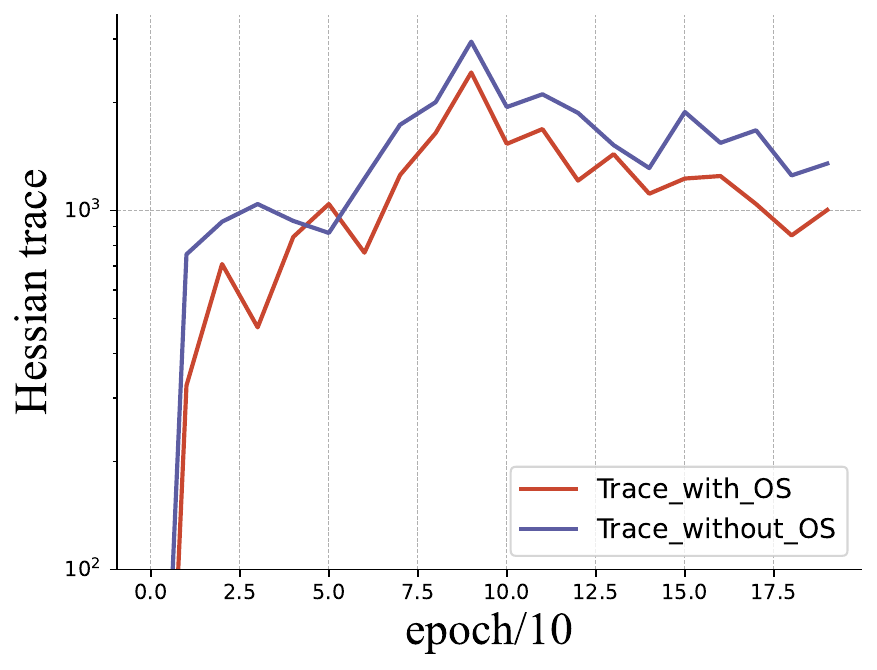}
    \subcaption{DenseNet201}
    \end{subfigure}
    \caption{Visualization of the Hessian trace for different models and dataset with and without SOS during training.
    }
    \label{Hessian during training}
\end{figure}

\noindent\textbf{Generalization Ability Test.}
Next, we demonstrate that SOS can also improve the generalization ability of neural networks when applied in the training process.
We apply SGD to train the same four CNN models under the CIFAR-10 and CIFAR-100 datasets with the same data augmentation strategies.
Following \cite{dense}, the weight decay is $10^{-4}$ and a Nesterov momentum of $0.9$ without damping.
The batch size is set to be $64$ and the models are trained for $300$ epochs.  
The initial learning rate is set to be $0.1$ and divided by 10  and is divided by 10 at 50\% and
75\% of the total number of training epochs.
All images are applied with a simple random horizontal flip and normalized using their mean and standard deviation. 
We focus on the comparisons between four different training schemes, namely the standard SGD scheme, SOS on SGD schemes, SAM scheme~\cite{SAM}, and SOS on SAM schemes.
Given that both SGD and SAM search optimal parameters within local regions, our SOS operates globally and thus can be integrated with SGD and SAM to attain better generalization.
%

As we can see in Table~\ref{accuray_table}, all the accuracy of the four CNN network architectures on CIFAR-10 and CIFAR-100 have been improved with SOS compared to the standard SGD and SAM schemes without SOS.
The test accuracy is also increased, for example, the test accuracy of ResNet-50 on CIFAR-100 has been improved from $81.65_{\pm 0.27}\%$ to $82.41_{\pm 0.33}\%$.
%

\noindent\textbf{Hessian Analysis.}
The evolution of the trace of Hessian during training with and without SOS is shown in Figure~\ref{Hessian during training}. We train VGG16 and ResNet18 on CIFAR10 and DenseNet201 on CIFAR100 dataset for 200 epochs and visualize the trace of Hessian during training. It shows that the trace of Hessian during training keeps increasing. However, when we apply SOS to the model during training, the trace of Hessian is minimized, which indicates a better generalization ability.

\begin{figure}[t]
    \centering
    \begin{minipage}{.45\textwidth}
            \begin{subfigure}{0.49\textwidth}
                \hspace{-.6cm}\includegraphics[width=1\linewidth]{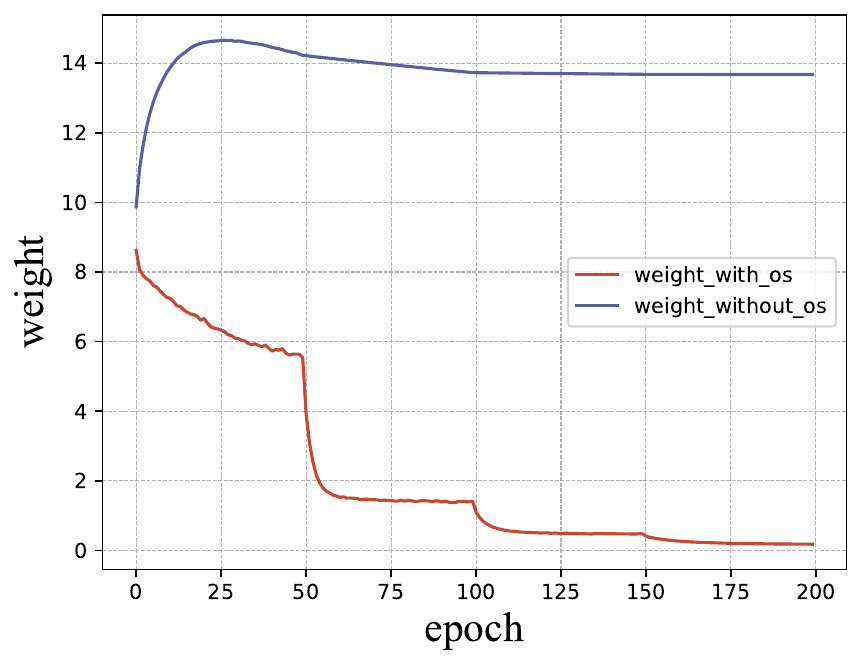}
                \caption*{\hspace{-.7cm} VGG w/ mixup}           
            \end{subfigure}  
            \begin{subfigure}{0.49\textwidth}
                \hspace{-.6cm}\includegraphics[width=1\linewidth]{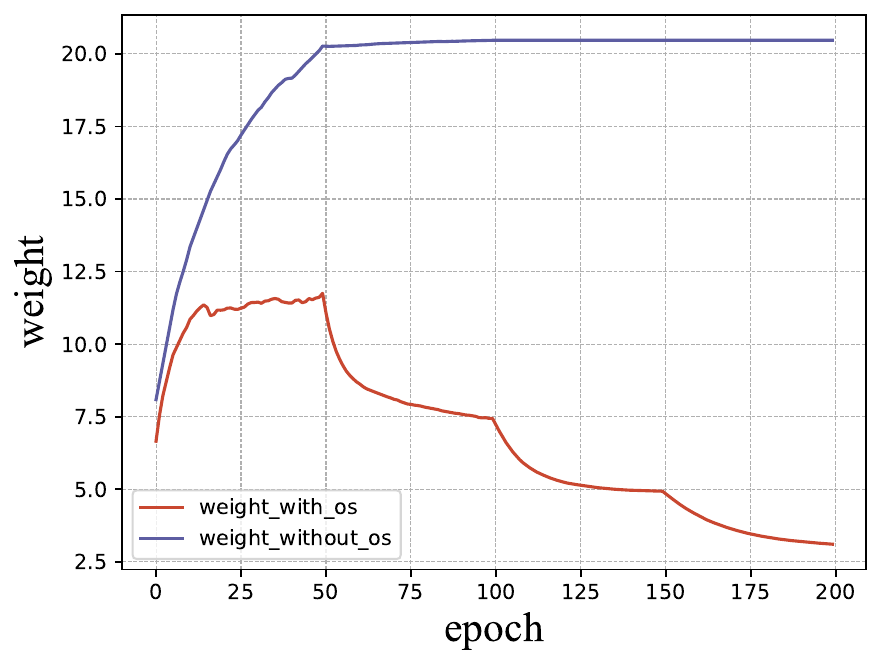}   
                \caption*{\hspace{-.9cm} DenseNet w/ mixup}             
            \end{subfigure}            
            \captionof{figure}{Weight visualization.}\label{fig3}
    \end{minipage}
    \begin{minipage}{0.5\textwidth}
        \centering
          \captionof{table}{ Object detection.}\label{tab4}
            \vskip 0.15in
        \begin{tabular}{c|c|c}
        \toprule
        VOC & Yolo-V5s & Yolo-V5x \\
        \midrule
        mAP w/ SOS	 & $83.4_{\pm 0.2}$ & $87.1_{\pm 0.2}$ \\ \midrule
        mAP w/o SOS	 & $\mathbf{83.7}_{\pm 0.1}$ & $\mathbf{87.4}_{\pm 0.2}$\\
        \bottomrule
        \end{tabular}
    \end{minipage}
\end{figure}

\noindent\textbf{Analysis on the Last-Layer Weight.}
The model parameters' weight with and without SOS is shown in Figure ~\ref{fig3}.
The Frobenius norm of the last linear layer's weight increases a lot before dividing the learning rate by 10.
It has been slowed down when the learning rate is divided and the weight starts to decrease rapidly for the model with OS.
When SOS is applied, the increase rate is slowed down for the DenseNet. For VGG, the weight does not increase but decreases from the first epoch.
%
%

\noindent\textbf{Discussion on SOS Batch Size.}
The SOS algorithm we proposed to satisfy Assumption~\ref{a1} introduces a hyper-parameter batch size, which has a significant impact on the performance of the algorithm.
For SOS batch size, there is a trade-off.
When the batch size is too large, it will limit the degrees of freedom for SOS to solve $ {A} {V}= {Z}$.
However, when the batch size is too small, it may not keep the training loss value unchanged. 
We conduct experiments to validate our statement on CIFAR100 dataset and show in Table~\ref{batchsize}.
We use different SOS batch sizes for training and report the accuracy.
As Table~\ref{batchsize} shows, with the increase in batch size, the improvement increases initially and then decreases. 
We can see that with $batchsize = 300$, it attains the best performance. 
\begin{table*}[t]
\caption{Investigation of SOS batch size on CIFAR100}
\label{sosb}
    \vskip 0.15in\hspace{-0.5cm} 
\resizebox{1.0\textwidth}{!}{
\setlength{\tabcolsep}{6.0mm}{
\renewcommand{\arraystretch}{1.2}
    \begin{tabular}{l|cccc}
    \toprule
        SOS batch size & \multicolumn{1}{|c}{w/o SOS} & \multicolumn{1}{c}{batch = 200} & \multicolumn{1}{c}{batch = 300} & \multicolumn{1}{c}{batch = 400} \\ \midrule
        VGG-16        & $\ \text{72.12}_{\pm \text{0.52}}\ $ & $\ \text{72.91}_{\pm 0.19}\ $ & $\ \textbf{73.35}_{\pm 0.43}\ $  & $\  \text{73.18}_{\pm 0.43}\ $ \\
        \midrule
        ResNet-18   & $\ \text{78.03}_{\pm 0.41}\ $ & $\ \text{79.23}_{\pm 0.19}\ $ & $\ \textbf{78.43}_{\pm 0.24}\ $  & $\ \text{95.97}_{\pm 0.21}\ $  \\
        \midrule
        ResNet-50   &  $\ \text{79.30}_{\pm 0.22}\ $ & $\ \text{79.53}_{\pm 0.21}\  $ & $\ \textbf{79.63}_{\pm 0.27}\   $  & $\ \text{79.30}_{\pm 0.22}\  $  \\
        \midrule
        DenseNet-201    & $\ \text{80.02}_{\pm 0.32}\ $ & $\ \text{80.27}_{\pm 0.23}\ $ & $\ \textbf{80.43}_{\pm 0.13}\ $  & $\ \text{80.33}_{\pm 0.18}\ $  \\
        \bottomrule
    \end{tabular}
    \label{batchsize}
    }
    }
\end{table*}
\noindent\textbf{Object Detection.}
Our method improves generalization performance on other recognition tasks.
We conduct experiments on other computer vision tasks, such as object detection to validate the improved generalization performance. 
Table \ref{tab4} shows the object detection baseline results with and without SOS on PASCAL VOC dataset~\cite{pascalvoc}.
We adopt YOLOv5s and YOLOv5x~\cite{yolov5} as the detection model and we see that the performance (mAP) of the two models is improved when applied with SOS.
\section{Complexity Analysis}
The time complexity for SOS contains two parts: Gaussian elimination and computing the inverse matrix.
For Gaussian Elimination, the complexity is $\mathcal{O}(b^2(m+n))$ and for computing the inverse matrix, the complexity is $\mathcal{O}(n^3)$, where $n$ and $m$ is the rows and columns of the weight matrx, $b$ is the batch size for SOS. 
Therefore the complexity for once SOS is
$\mathcal{O}(b^2(m+n)+n^3)$.
If applied during training in each epoch, the whole time complexity is $\mathcal{O}(epoch(b^2(m+n)+m^3))$. Here $b$ is the batch size and $m, n$ are the number of rows and columns of the weight matrix.

\section{Conclusion}
In this paper, we introduce a novel technique called optimum shifting, to move the parameters of neural networks from sharper minima to flatter minima while keeping the training loss value unchanged.
It treats the matrix multiplications in the network as systems of under-determined linear equations and modifies parameters in the solution space.
To reduce the computational costs and increase the degrees of freedom for optimum shifting, we proposed stochastic optimum shifting, which selects a small batch for optimum shifting. 
The neural collapse phenomenon guarantees that the proposed stochastic optimum shifting remains empirical loss unchanged.
We perform experiments to show that stochastic optimum shifting keeps the empirical loss unchanged, reduces the Hessian trace, and improves the generalization ability on different vision tasks.

\newpage

\bibliography{example_paper}

\begin{thebibliography}{10}\itemsep=-1pt

\bibitem{blanc2020implicit}
Guy Blanc, Neha Gupta, Gregory Valiant, and Paul Valiant.
\newblock Implicit regularization for deep neural networks driven by an ornstein-uhlenbeck like process.
\newblock In {\em Conference on learning theory}, pages 483--513. PMLR, 2020.

\bibitem{chaudhari2019entropy}
Pratik Chaudhari, Anna Choromanska, Stefano Soatto, Yann LeCun, Carlo Baldassi, Christian Borgs, Jennifer Chayes, Levent Sagun, and Riccardo Zecchina.
\newblock {Entropy-SGD}: Biasing gradient descent into wide valleys.
\newblock {\em Journal of Statistical Mechanics: Theory and Experiment}, 2019(12):124018, 2019.

\bibitem{dinh}
Laurent Dinh, Razvan Pascanu, Samy Bengio, and Yoshua Bengio.
\newblock Sharp minima can generalize for deep nets.
\newblock In {\em ICML}, pages 1019--1028. PMLR, 2017.

\bibitem{pascalvoc}
Mark Everingham, Luc~Van Gool, Christopher K.~I. Williams, John~M. Winn, and Andrew Zisserman.
\newblock The pascal visual object classes (voc) challenge.
\newblock {\em Int. J. Comput. Vis.}, 88(2):303--338, 2010.

\bibitem{SAM}
Pierre Foret, Ariel Kleiner, Hossein Mobahi, and Behnam Neyshabur.
\newblock Sharpness-aware minimization for efficiently improving generalization.
\newblock {\em ICLR}, 2021.

\bibitem{gatmiry2023inductive}
Khashayar Gatmiry, Zhiyuan Li, Ching-Yao Chuang, Sashank Reddi, Tengyu Ma, and Stefanie Jegelka.
\newblock The inductive bias of flatness regularization for deep matrix factorization.
\newblock {\em arXiv preprint arXiv:2306.13239}, 2023.

\bibitem{gans}
Ian Goodfellow, Jean Pouget-Abadie, Mehdi Mirza, Bing Xu, David Warde-Farley, Sherjil Ozair, Aaron Courville, and Yoshua Bengio.
\newblock Generative adversarial networks.
\newblock {\em Communications of the ACM}, 63(11):139--144, 2020.

\bibitem{resnet}
Kaiming He, Xiangyu Zhang, Shaoqing Ren, and Jian Sun.
\newblock Deep residual learning for image recognition.
\newblock In {\em CVPR}, 2016.

\bibitem{hochreiter1994simplifying}
Sepp Hochreiter and J{\"u}rgen Schmidhuber.
\newblock Simplifying neural nets by discovering flat minima.
\newblock {\em NeurIPS}, 7, 1994.

\bibitem{dense}
Gao Huang, Zhuang Liu, Laurens Van Der~Maaten, and Kilian~Q Weinberger.
\newblock Densely connected convolutional networks.
\newblock In {\em CVPR}, 2017.

\bibitem{jiang2019fantastic}
Yiding Jiang, Behnam Neyshabur, Hossein Mobahi, Dilip Krishnan, and Samy Bengio.
\newblock Fantastic generalization measures and where to find them.
\newblock {\em arXiv preprint arXiv:1912.02178}, 2019.

\bibitem{yolov5}
Glenn Jocher, Alex Stoken, Jirka Borovec, NanoCode012, ChristopherSTAN, Liu Changyu, Laughing, tkianai, Adam Hogan, lorenzomammana, yxNONG, AlexWang1900, Laurentiu Diaconu, Marc, wanghaoyang0106, ml5ah, Doug, Francisco Ingham, Frederik, Guilhen, Hatovix, Jake Poznanski, Jiacong Fang, Lijun Yu, changyu98, Mingyu Wang, Naman Gupta, Osama Akhtar, PetrDvoracek, and Prashant Rai.
\newblock {ultralytics/yolov5: v3.1 - Bug Fixes and Performance Improvements}, Oct. 2020.

\bibitem{minima}
Nitish~Shirish Keskar, Dheevatsa Mudigere, Jorge Nocedal, Mikhail Smelyanskiy, and Ping Tak~Peter Tang.
\newblock On large-batch training for deep learning: Generalization gap and sharp minima.
\newblock {\em arXiv preprint arXiv:1609.04836}, 2016.

\bibitem{keskar2017large}
Nitish~Shirish Keskar, Dheevatsa Mudigere, Jorge Nocedal, Mikhail Smelyanskiy, and Ping Tak~Peter Tang.
\newblock On large-batch training for deep learning: Generalization gap and sharp minima.
\newblock In {\em ICLR}, 2017.

\bibitem{vae}
Diederik~P Kingma and Max Welling.
\newblock Auto-encoding variational {Bayes}.
\newblock {\em arXiv preprint arXiv:1312.6114}, 2013.

\bibitem{CIFAR10}
Alex Krizhevsky, Geoffrey Hinton, et~al.
\newblock Learning multiple layers of features from tiny images.
\newblock 2009.

\bibitem{asam}
Jungmin Kwon, Jeongseop Kim, Hyunseo Park, and In~Kwon Choi.
\newblock Asam: Adaptive sharpness-aware minimization for scale-invariant learning of deep neural networks.
\newblock In {\em ICML}. PMLR, 2021.

\bibitem{li2018visualizing}
Hao Li, Zheng Xu, Gavin Taylor, Christoph Studer, and Tom Goldstein.
\newblock Visualizing the loss landscape of neural nets.
\newblock {\em NeurIPS}, 2018.

\bibitem{neyshabur2017implicit}
Behnam Neyshabur.
\newblock Implicit regularization in deep learning.
\newblock {\em arXiv preprint arXiv:1709.01953}, 2017.

\bibitem{gpt4}
OpenAI.
\newblock {GPT-4} technical report, 2023.

\bibitem{vgg}
Karen Simonyan and Andrew Zisserman.
\newblock Very deep convolutional networks for large-scale image recognition.
\newblock {\em arXiv preprint arXiv:1409.1556}, 2014.

\bibitem{dpm}
Jascha Sohl-Dickstein, Eric Weiss, Niru Maheswaranathan, and Surya Ganguli.
\newblock Deep unsupervised learning using nonequilibrium thermodynamics.
\newblock In {\em ICML}, pages 2256--2265. PMLR, 2015.

\bibitem{tirer2022extended}
Tom Tirer and Joan Bruna.
\newblock Extended unconstrained features model for exploring deep neural collapse.
\newblock In {\em ICML}, pages 21478--21505. PMLR, 2022.

\bibitem{transformer}
Ashish Vaswani, Noam Shazeer, Niki Parmar, Jakob Uszkoreit, Llion Jones, Aidan~N Gomez, {\L}ukasz Kaiser, and Illia Polosukhin.
\newblock Attention is all you need.
\newblock {\em NeurIPS}, 2017.

\bibitem{matlab}
Andrea Vedaldi and Karel Lenc.
\newblock Matconvnet: Convolutional neural networks for matlab.
\newblock In {\em ACM MM}, pages 689--692, 2015.

\bibitem{wen2023sharpness}
Kaiyue Wen, Tengyu Ma, and Zhiyuan Li.
\newblock Sharpness minimization algorithms do not only minimize sharpness to achieve better generalization.
\newblock {\em arXiv preprint arXiv:2307.11007}, 2023.

\bibitem{jxwu}
Jianxin Wu.
\newblock {\em Essentials of pattern recognition: an accessible approach}.
\newblock Cambridge University Press, 2020.

\bibitem{pyhessain}
Zhewei Yao, Amir Gholami, Kurt Keutzer, and Michael~W Mahoney.
\newblock Pyhessian: Neural networks through the lens of the {Hessian}.
\newblock In {\em 2020 IEEE international conference on big data (Big data)}, pages 581--590. IEEE, 2020.

\bibitem{mixup}
Hongyi Zhang, Moustapha Cisse, Yann~N Dauphin, and David Lopez-Paz.
\newblock Mixup: Beyond empirical risk minimization.
\newblock {\em arXiv preprint arXiv:1710.09412}, 2017.

\bibitem{zhaopenalizing}
Yang Zhao, Hao Zhang, and Xiuyuan Hu.
\newblock Penalizing gradient norm for efficiently improving generalization in deep learning.
\newblock In {\em ICML}. PMLR, 2022.

\bibitem{zhou2022optimization}
Jinxin Zhou, Xiao Li, Tianyu Ding, Chong You, Qing Qu, and Zhihui Zhu.
\newblock On the optimization landscape of neural collapse under {MSE} loss: Global optimality with unconstrained features.
\newblock In {\em ICML}. PMLR, 2022.

\bibitem{nc2}
Zhihui Zhu, Tianyu Ding, Jinxin Zhou, Xiao Li, Chong You, Jeremias Sulam, and Qing Qu.
\newblock A geometric analysis of neural collapse with unconstrained features.
\newblock {\em NeurIPS}, 2021.

\end{thebibliography}
\bibliographystyle{nips}

\newpage
\appendix
\onecolumn
\section{Appendix}
\label{appendix a}
\subsection{Proof of Theorem 3.2. with CNN}
\label{proof of theorem 1}
We denote $\boldsymbol{x}_{l, i}$ as the output of $l$-th layer of neural network with input data sample $\boldsymbol{x}_i$:
\begin{align}
    \boldsymbol{x}_{l, i} = \sigma(\boldsymbol{F}_{l}\ast \cdots \sigma (\boldsymbol{F}_1\ast\boldsymbol{x}_{i})\cdots   ).
\end{align}
The Hessian trace of loss function $\text{tr}(\mathbf{H}_{L})$ can be represented as:
\begin{align}
	\text{tr}(\mathbf{H}_{L}) =\sum_{p=1}^{m_{L+1}} \text{tr}(\nabla^2_{\bm{V}_p}L ) + \sum_{l_0=1}^l \text{tr}(\nabla^2_{\text{vec}(\boldsymbol{F}_{l_0})}L ),
\end{align}
where $\bm{V}_p$ denotes the $p$-th column of $\bm{V}$. The gradient of $L$ with respect to $\bm{V}$ is
\begin{align}
    \frac{\partial L}{\partial \bm{V}_p} 
    &= \frac{1}{n} \sum_{i=1}^n \frac{\partial L(f(\boldsymbol{x}_i), \boldsymbol{y}_i)}{\partial f_p(\boldsymbol{x}_i)^T}\frac{\partial f_p(\boldsymbol{x}_i)}{\partial \bm{V}_p} =\frac{1}{n} \sum_{i=1}^n \frac{\partial  L(f(\boldsymbol{x}_i), \boldsymbol{y}_i)}{\partial f_p(\boldsymbol{x}_i)^T}\textup{vec}(\boldsymbol{x}_{L, i})^T,
\end{align}
where $f_p(\boldsymbol{x}_i)$ denotes the $p$-th element in $f(\boldsymbol{x}_i)$.
So, the second-order derivative is as follows:
\begin{align}
    \nabla_{\bm{V}_p}^2L = \frac{1}{n}\sum_{i=1}^n  \frac{\partial^2 L(f(\boldsymbol{x}_i), \boldsymbol{y}_i)}{\partial f_p(\boldsymbol{x}_i)^T \partial f_p(\boldsymbol{x}_i)^T}\textup{vec}(\boldsymbol{x}_{L, i})\textup{vec}(\boldsymbol{x}_{L, i})^T.
\end{align}
Because the output of the classifier does not change, $\nabla_{\bm{V}}^2L$ is independent of $\bm{V}$ and will not change. For the second part:
\begin{align}
    \nabla_{\text{vec}(\boldsymbol{F}_{l_0})} L 
    &=\frac{1}{n} \sum_{i=1}^n \frac{\partial  L(f(\boldsymbol{x}_i), \boldsymbol{y}_i)}{\partial f(\boldsymbol{x}_i)^T}\left\{ \bm{V}^T  \prod_{q=l+1}^{L-1}\left[ ( \Sigma_{q, i}\textup{vec}(\boldsymbol{F}_q)^T\otimes I_q ) M_q\right]\left[\Sigma_{l, i} I_{l} \otimes \phi(\boldsymbol{x}_{l, i})\right]\right\}, 
\end{align}
where $\Sigma_{l, i} = \text{Diag}\left[ \mathbbm{1}\left\{ \text{vec}(\boldsymbol{x}_{l, i})> 0 \right\} \right]$, $I_l\in \mathbb{R}^{|\boldsymbol{x}_{l+1, i}|\times|\boldsymbol{x}_{l+1, i}|}$\footnote{$|A|$ denotes the number of elements in $A$.} is an identity matrix, and $M_l$ is an indicator matrix satisfying $\text{vec}(\phi(\boldsymbol{x}_{l, i})) = M_l\text{vec}(\boldsymbol{x}_{l, i})$~\cite{jxwu, matlab}. So the second order derivative is:
\begin{align}
\nabla^2_{\text{vec}(\boldsymbol{F}_{l_0})}L
    = &\frac{1}{n}\sum_{i=1}^n  \frac{\partial^2  L(f(\boldsymbol{x}_i), \boldsymbol{y}_i)}{\partial f(\boldsymbol{x}_i)^T \partial f(\boldsymbol{x}_i)^T}\left\{
    \bm{V}^T  \prod_{q=l+1}^{L-1}\left[ ( \Sigma_{q, i}\textup{vec}( \boldsymbol{F}_q)^T\otimes I_q ) M_q\right]\left[\Sigma_{l, i} I_{l} \otimes \phi(\boldsymbol{x}_{l, i})\right]\right\}^T \cdot  \notag \\ 
    &\left\{ \bm{V}^T  \prod_{q=l+1}^{L-1}\left[ ( \Sigma_{q, i} \textup{vec}(\boldsymbol{F}_q)^T\otimes I_q ) M_q\right]\left[\Sigma_{l, i} I_{l} \otimes \phi(\boldsymbol{x}_{l, i})\right]\right\}. 
\end{align}
We use $\mathcal{S}$ to denote $\prod_{q=l+1}^{L-1}\left[ ( \Sigma_{q, i} \textup{vec}(\boldsymbol{F}_q)^T\otimes I_q ) M_q\right]\left[\Sigma_{l, i} I_{l} \otimes \phi(\boldsymbol{x}_{l, i})\right]$ . And tr$(H_L)$ can be expressed as:
\begin{align}
    \text{tr}(\mathbf{H}_L) =&  \frac{1 }{n}\sum_{i=1}^n \sum_{p=1}^{m_{L+1}} 
    \frac{\partial^2 L(f(\boldsymbol{x}_i), \boldsymbol{y}_i)}{\partial f_p(\boldsymbol{x}_i)^T\partial f_p(\boldsymbol{x}_i)^T }\Vert\textup{vec}(\boldsymbol{x}_{L, i})\Vert^2 + \frac{1}{n}\sum_{l_0=1}^l \sum_{i=1}^n  \text{tr}\left\{ \frac{\partial^2  L(f(\boldsymbol{x}_i), \boldsymbol{y}_i)}{\partial f(\boldsymbol{x}_i)^T \partial f(\boldsymbol{x}_i)^T} 
    \mathcal{S}^T \bm{V}
    \bm{V}^T  \mathcal{S}\right\}
    \label{trace1}
\end{align}
So the trace of Hessian can be upper bounded by:
\begin{align}
    \text{tr}(\mathbf{H}_{L})
    \leq & \frac{1 }{n}\sum_{i=1}^n \sum_{p=1}^{m_{L+1}} 
    \frac{\partial^2 L(f(\boldsymbol{x}_i), \boldsymbol{y}_i)}{\partial f_p(\boldsymbol{x}_i)^T\partial f_p(\boldsymbol{x}_i)^T }\Vert\textup{vec}(\boldsymbol{x}_{L, i})\Vert^2 + 
    \frac{\Vert\bm{V}\Vert^2}{n}\sum_{l_0=1}^l \sum_{i=1}^n  \text{tr}\left\{ \frac{\partial^2  L(f(\boldsymbol{x}_i), \boldsymbol{y}_i)}{\partial f(\boldsymbol{x}_i)^T \partial f(\boldsymbol{x}_i)^T} 
    \mathcal{S}^T \mathcal{S}\right\},
\end{align}
Thus, we have finished the proof of the upper bound. For the lower bound, because $\mathcal{SS}^T$ is semi-positive definite, so we factorized it as:
\begin{align}
    \mathcal{SS}^T = Q\begin{bmatrix}
     \Lambda & 0\\
     0  & 0
    \end{bmatrix}Q^T
\end{align}
where $Q$ is orthogonal matrix and $\Lambda$ is is a diagonal matrix. The second part of \cref{trace1} can be lower bounded as:
\begin{align}
    \nabla^2_{\text{vec}(\boldsymbol{F}_l)}L 
    &= \frac{1}{n}\sum_{l_0=1}^l\sum_{i=1}^n  \frac{\partial^2  L(f(\boldsymbol{x}_i), \boldsymbol{y}_i)}{\partial f(\boldsymbol{x}_i)^T \partial f(\boldsymbol{x}_i)^T} \text{tr}( 
    \mathcal{S}^T \bm{V}
    \bm{V}^T  \mathcal{S})\\
    &= \frac{1}{n}\sum_{l_0=1}^l\sum_{i=1}^n  \frac{\partial^2  L(f(\boldsymbol{x}_i), \boldsymbol{y}_i)}{\partial f(\boldsymbol{x}_i)^T \partial f(\boldsymbol{x}_i)^T} \text{tr}( 
     \bm{V}
    \bm{V}^T  Q\begin{bmatrix}
     \Lambda & 0\\
     0  & 0
    \end{bmatrix}Q^T )\\
    &= \frac{1}{n}\sum_{l_0=1}^l\sum_{i=1}^n \frac{\partial^2  L(f(\boldsymbol{x}_i), \boldsymbol{y}_i)}{\partial f(\boldsymbol{x}_i)^T \partial f(\boldsymbol{x}_i)^T}  \frac{\text{tr}( 
     \bm{V}
    \bm{V}^T  Q\begin{bmatrix}
     \Lambda & 0\\
     0  & 0
    \end{bmatrix}Q^T )\text{tr}(\begin{bmatrix}
     \Lambda^{-1} & 0\\
     0  & 0
    \end{bmatrix})}{\text{tr}(\begin{bmatrix}
     \Lambda^{-1} & 0\\
     0  & 0
    \end{bmatrix})}\\
    &\geq \frac{1}{n}\sum_{l_0=1}^l\sum_{i=1}^n \frac{\partial^2  L(f(\boldsymbol{x}_i), \boldsymbol{y}_i)}{\partial f(\boldsymbol{x}_i)^T \partial f(\boldsymbol{x}_i)^T}  \frac{\Vert
     \bm{V}\Vert^2}{\text{tr}(\Lambda^{-1})}\\
\end{align}
Thus, we have finished the proof of lower bound.

The first part is independent with $\Vert\bm{V}\Vert^2$. And the second part is linearly dependent on  $\Vert\bm{V}\Vert^2$. So if $\Vert\bm{V}\Vert^2$ is minimized, then the upper and lower bound of the Hessian trace will also be minimized, and we end the proof.

\subsection{Proof of Theorem 3.2. with ResNet}
\label{proof of theorem 2}
For ResNet, which uses skip connection that bypasses the non-linear transformations:
\begin{align}
    \boldsymbol{x}_{2k+1, i} 
    &= \sigma(\boldsymbol{F}_{2k} \ast \sigma( \boldsymbol{F}_{2k-1} \ast \boldsymbol{x}_{2k-1, i} ) +\boldsymbol{x}_{2k-1, i}  )\\
    &= \sigma(\boldsymbol{F}_{2k} \ast \boldsymbol{x}_{2k, i} +\boldsymbol{x}_{2k-1, i}  )
\end{align}
It can also be represented as:
\begin{align}
    \text{vec}(\boldsymbol{x}_{2k+1, i}) 
    &= \Sigma_{2k, i} \left[ \text{vec}( \boldsymbol{F}_{2k} \ast \boldsymbol{x}_{2k, i}) + \text{vec}(\boldsymbol{x}_{2k-1, i}) \right]\\
    &= \left\{\begin{matrix} 
  \Sigma_{2k, i} I_{2k}\otimes\phi(\boldsymbol{x}_{2k, i})\text{vec}(\boldsymbol{F}_{2k}) + \Sigma_{2k, i}\text{vec}(\boldsymbol{x}_{2k-1, i}) \\  \\
  \Sigma_{2k, i} (\boldsymbol{F}_{2k}^T\otimes I_{2k})M_{2k}\text{vec}(\boldsymbol{x}_{2k, i}) + \Sigma_{2k, i} \text{vec}( \boldsymbol{x}_{2k-1, i} )
\end{matrix}\right.
\end{align}
So the gradient with respect to vec$(\boldsymbol{F})$ and vec$(\boldsymbol{x}_{2k-1})$ is:
\begin{align}
    \frac{\partial \text{vec} (\boldsymbol{x}_{2k+1, i})}{\partial \text{vec}(\boldsymbol{x}_{2k-1, i})^T} = \Sigma_{2k, i} (\boldsymbol{F}_{2k}^T\otimes I_{2k})M_{2k}\Sigma_{2k-1, i} (\boldsymbol{F}_{2k-1}^T\otimes I_{2k-1})M_{2k-1} + \Sigma_{2k, i}
\end{align}
\begin{align}
    \frac{\partial \text{vec} (\boldsymbol{x}_{2k+1, i})}{\partial \text{vec}(\boldsymbol{F}_{2k})^T} = \Sigma_{2k, i} I_{2k}\otimes\phi(\boldsymbol{x}_{2k, i})
\end{align}
\begin{align}
    \frac{\partial \text{vec} (\boldsymbol{x}_{2k+1, i})}{\partial \text{vec}(\boldsymbol{F}_{2k-1})^T} = \Sigma_{2k, i} (\boldsymbol{F}_{2k}^T\otimes I_{2k})M_{2k}  \Sigma_{2k-1, i} I_{2k-1}\otimes\phi(\boldsymbol{x}_{2k-1, i})
\end{align}
We assume that $l$ is multiples of $2$, i.e. $\exists t$ s.t. $l = 2t$, and we represent the neural network as follows:
\begin{align}
    f(\boldsymbol{x}_i) = \bm{V}^T \text{vec}( \sigma (\boldsymbol{F}_{2t}\ast \boldsymbol{x}_{2t, i} + \boldsymbol{x}_{2t-1, i}) )
\end{align}
The gradient of $f$ with respect to vec$(\boldsymbol{F}_{2l_0})$ and vec$(\boldsymbol{F}_{2l_0-1})$ are:
\begin{align}
    \nabla_{\text{vec}(\boldsymbol{F}_{2l_0})} f 
    &= \frac{\partial f}{\partial \text{vec} ( \boldsymbol{x}_{2t+1, i})^T } 
    \left[\prod_{r=l+1}^t (\frac{\partial \text{vec}(\boldsymbol{x}_{2r+1, i})}{\partial \text{vec}(\boldsymbol{x}_{2r-1, i})^T})\right]  \frac{\partial \text{vec}(\boldsymbol{x}_{2l_0+1, i})}{\partial\text{vec}(\boldsymbol{F}_{2l})^T}\\
    \nabla_{\text{vec}(\boldsymbol{F}_{2l_0-1})} f &= \frac{\partial f}{\partial \text{vec}(\boldsymbol{x}_{2t+1, i})^T} \left[
    \prod_{r=l+1}^t (\frac{\partial\text{vec}(\boldsymbol{x}_{2r+1, i})}{\partial \text{vec}(\boldsymbol{x}_{2r-1, i})^T})\right]  \frac{\partial \text{vec}(\boldsymbol{x}_{2l_0+1, i})}{\partial \text{vec}(\boldsymbol{F}_{2l-1})^T}\\
\end{align}
We denote $\left[\prod_{r=l+1}^{L-1} (\frac{\partial \boldsymbol{x}_{2r+1, i}}{\partial \boldsymbol{x}_{2r-1, i}^T})   \right]\left[\mathbbm{1}_{r\%2 = 0}\left[\frac{\partial \text{vec}(\boldsymbol{x}_{2l_0+1, i})}{\partial\text{vec}(\boldsymbol{F}_{2l})^T}\right] + \mathbbm{1}_{r\%2 = 1}\left[\frac{\partial \text{vec}(\boldsymbol{x}_{2l_0+1, i})}{\partial\text{vec}(\boldsymbol{F}_{2l-1})^T}\right] \right]$ as $\mathcal{S}$. The Hessian trace can be represented as
\begin{align}
    \text{tr}(H_{L}) 
    =& \frac{1}{n}\sum_{q=1}^{m_{L+1}}  \sum_{i=1}^n \frac{\partial^2 L(f(\boldsymbol{x}_i), \boldsymbol{y}_i)}{\partial f_q(\boldsymbol{x}_i)^T\partial f_q(\boldsymbol{x}_i)^T}\Vert \text{vec}(\boldsymbol{x}_{L, i})\Vert^2 +\\
    &\frac{1}{n} \sum_{i=1}^n \sum_{l_0=1}^{l}\text{tr}\left\{ \frac{\partial^2 L(f(\boldsymbol{x}_i), \boldsymbol{y}_i)}{\partial f(\boldsymbol{x}_i)^T\partial f(\boldsymbol{x}_i)^T } \mathcal{S}^T \bm{V} \bm{V}^TS \right\}
\end{align}
This has the same form as \cref{trace1}. So the Hessian trace is upper bounded by:
\begin{align}
    \text{tr}(\mathbf{H}_{L})
    \leq & \frac{1 }{n}\sum_{i=1}^n \sum_{p=1}^{m_{L+1}} 
    \frac{\partial^2 L(f(\boldsymbol{x}_i), \boldsymbol{y}_i)}{\partial f_p(\boldsymbol{x}_i)^T\partial f_p(\boldsymbol{x}_i)^T }\Vert\textup{vec}(\boldsymbol{x}_{L, i})\Vert^2 + 
    \frac{\Vert\bm{V}\Vert^2}{n}\sum_{l_0=1}^l \sum_{i=1}^n  \text{tr}\left\{ \frac{\partial^2  L(f(\boldsymbol{x}_i), \boldsymbol{y}_i)}{\partial f(\boldsymbol{x}_i)^T \partial f(\boldsymbol{x}_i)^T} 
    \mathcal{S}^T \mathcal{S}\right\},
\end{align}
and lower bounded by:
\begin{align}
    \nabla^2_{\text{vec}(\boldsymbol{F}_l)}L 
    &\geq \frac{1 }{n}\sum_{i=1}^n \sum_{p=1}^{m_{L+1}} 
    \frac{\partial^2 L(f(\boldsymbol{x}_i), \boldsymbol{y}_i)}{\partial f_p(\boldsymbol{x}_i)^T\partial f_p(\boldsymbol{x}_i)^T }\Vert\textup{vec}(\boldsymbol{x}_{L, i})\Vert^2 + \frac{1}{n}\sum_{l_0=1}^l\sum_{i=1}^n \frac{\partial^2  L(f(\boldsymbol{x}_i), \boldsymbol{y}_i)}{\partial f(\boldsymbol{x}_i)^T \partial f(\boldsymbol{x}_i)^T}  \frac{\Vert
     \bm{V}\Vert^2}{\text{tr}(\Lambda^{-1})}
\end{align}

\subsection{Proof of Proof of Theorem 3.2. with DenseNet}
\label{proof of theorem 3}
When neural networks using dense connective pattern, which is represented as follows:
\begin{align}
    \boldsymbol{x}_{l+1, i}
    &= \sigma(\boldsymbol{F}_l\ast \left[\boldsymbol{x}_{l, i}, \cdots,  \boldsymbol{x}_{1, i} \right]) \\
    & = \sigma( \sum_{j=1}^l \boldsymbol{F}_l\ast\boldsymbol{x}_{j, i})
\end{align}
We vectorize it and represented the dense connection as:
\begin{align}
    \text{vec}(\boldsymbol{x}_{l+1, i}) & = \Sigma_{l, i} \left[\sum_{j=1}^l (\boldsymbol{F}_{j}^T\otimes I_{j})\text{vec}(\phi(\boldsymbol{x}_{j, i})) \right]\\
    &= \sum_{j=1}^l  \left[\Sigma_{l, i}  (\boldsymbol{F}_{j}^T\otimes I_{j})\text{vec}(\phi(\boldsymbol{x}_{j, i}))   \right]
\end{align}
The neural network with $L$ convolutional layer can be represented as:
\begin{align}
    f(\boldsymbol{x}_i) = \bm{V}^T \text{vec}(\boldsymbol{x}_{L, i})
\end{align}
The gradient of $p$-th layer output with respect to $q$-th layer output is:
\begin{align}
    \frac{\partial \text{vec}(\boldsymbol{x}_{p, i})}{\partial\text{vec} (\boldsymbol{x}_{q, i})^T} =  \sum_{o=q}^{p} \prod_{j=q}^o \Sigma_{j, i} (\boldsymbol{F}_{j}^T\otimes I_{j})M_j 
\end{align}
So the gradient of $f$ with respect to vec$(\boldsymbol{F}_l)$ is:
\begin{align}
    \frac{\partial f(\boldsymbol{x}_i) }{\partial \text{vec}(\boldsymbol{F}_l)^T } 
    &= \frac{\partial f (\boldsymbol{x}_i) }{\partial \text{vec}(\boldsymbol{x}_{L+1})^T}\frac{\partial \text{vec}(\boldsymbol{x}_{L+1})}{\partial \text{vec}(\boldsymbol{x}_{l+1})^T}\frac{\partial \text{vec}(\boldsymbol{x}_{l+1})}{\partial \text{vec}(\boldsymbol{F}_{l})^T}\\
    &= \bm{V}^T  \left[\sum_{o=l+1}^{L} \prod_{j=q}^o \Sigma_{j, i} (\boldsymbol{F}_{j}^T\otimes I_{j})M_j\right] \Sigma_{l, i} I_{l}\otimes\phi(\boldsymbol{x}_{l, i}) \\
    &\coloneqq \bm{V}^T \mathcal{S}
\end{align}
The trace of Hessian can be represented as
\begin{align}
    \text{tr}(H_{L}) 
    =& \frac{1}{n}\sum_{q=1}^{m_{L+1}}  \sum_{i=1}^n \frac{\partial^2 L(f(\boldsymbol{x}_i), \boldsymbol{y}_i)}{\partial f_q(\boldsymbol{x}_i)^T\partial f_q(\boldsymbol{x}_i)^T}\Vert \text{vec}(\boldsymbol{x}_{L, i})\Vert^2 +\\
    &\frac{1}{n} \sum_{i=1}^n \sum_{l_0=1}^{l}\text{tr}\left\{ \frac{\partial^2 L(f(\boldsymbol{x}_i), \boldsymbol{y}_i)}{\partial f(\boldsymbol{x}_i)^T\partial f(\boldsymbol{x}_i)^T } \mathcal{S}^T \bm{V} \bm{V}^TS \right\}
\end{align}
This has the same form as \cref{trace1}. So the Hessian trace is upper bounded by:
\begin{align}
    \text{tr}(\mathbf{H}_{L})
    \leq & \frac{1 }{n}\sum_{i=1}^n \sum_{p=1}^{m_{L+1}} 
    \frac{\partial^2 L(f(\boldsymbol{x}_i), \boldsymbol{y}_i)}{\partial f_p(\boldsymbol{x}_i)^T\partial f_p(\boldsymbol{x}_i)^T }\Vert\textup{vec}(\boldsymbol{x}_{L, i})\Vert^2 + 
    \frac{\Vert\bm{V}\Vert^2}{n}\sum_{l_0=1}^l \sum_{i=1}^n  \text{tr}\left\{ \frac{\partial^2  L(f(\boldsymbol{x}_i), \boldsymbol{y}_i)}{\partial f(\boldsymbol{x}_i)^T \partial f(\boldsymbol{x}_i)^T} 
    \mathcal{S}^T \mathcal{S}\right\},
\end{align}
and lower bounded by:
\begin{align}
    \nabla^2_{\text{vec}(\boldsymbol{F}_l)}L 
    &\geq \frac{1 }{n}\sum_{i=1}^n \sum_{p=1}^{m_{L+1}} 
    \frac{\partial^2 L(f(\boldsymbol{x}_i), \boldsymbol{y}_i)}{\partial f_p(\boldsymbol{x}_i)^T\partial f_p(\boldsymbol{x}_i)^T }\Vert\textup{vec}(\boldsymbol{x}_{L, i})\Vert^2 + \frac{1}{n}\sum_{l_0=1}^l\sum_{i=1}^n \frac{\partial^2  L(f(\boldsymbol{x}_i), \boldsymbol{y}_i)}{\partial f(\boldsymbol{x}_i)^T \partial f(\boldsymbol{x}_i)^T}  \frac{\Vert
     \bm{V}\Vert^2}{\text{tr}(\Lambda^{-1})}
\end{align}

\subsection{Proof of Theorem 3.2. with MLP}
\label{proof of theorem 4}
For a fully connected neural network. Let $\{(\textbf{x}_1, \boldsymbol{y}_1), \cdots, (\textbf{x}_n, \boldsymbol{y}_n)\}$ be a set of n training samples. We consider l-hidden-layer neural networks as follows. 
\begin{align}
	f(\boldsymbol{x}) = \bm{V}^T\sigma(\boldsymbol{W}_l^T\sigma(\boldsymbol{W}_{l-1}^T\cdots \sigma(\boldsymbol{W}_1^T\boldsymbol{x})\cdots))
\end{align}
where $\sigma(x)=\max \{0, x\}$ is the entry-wise ReLU activation. $\boldsymbol{W}_l\in \mathbb{R}^{m_{l-1}\times m_l }$. Given input $\boldsymbol{x}_i$, We denote the output after the $l_0$-th layer using $\boldsymbol{x}_{l_0, i}$
\begin{align}
	\boldsymbol{x}_{l_0, i}
	&= \sigma (\boldsymbol{W}_{l_0}^T \sigma (\boldsymbol{W}_{l_0-1}^T\cdots \sigma (\boldsymbol{W}_1^T \boldsymbol{x}_i)\cdots))\\
	&= (\prod_{r=1}^{l} \Sigma_{r, i}\boldsymbol{W}_r^T )\boldsymbol{x}_{i}
\end{align}
where $\Sigma_{1, i} = \text{Diag}(\mathbbm{1}\{\boldsymbol{W}_1^T\boldsymbol{x}_i>0\})$, and $\Sigma_{l_0, i} = \text{Diag}\left[ \mathbbm{1}\left\{ \boldsymbol{W}_{l_0}^T (\prod_{r=1}^{l_0  -1}\Sigma_{i, r}\boldsymbol{W}_r^T)\boldsymbol{x}_i> 0 \right\} \right]$.
We have $f(\boldsymbol{x}_i) = \bm{V}^T\boldsymbol{x}_{l, i}$.
The Hessian trace of loss function $\text{tr}(\mathbf{H}_{L})$ can be represented as:
\begin{align}
	\text{tr}(\mathbf{H}_{L}) =\sum_{p=1}^{m_{l+1}} \text{tr}(\nabla^2_{\bm{V}_p}L ) + \sum_{l_0=1}^l \text{tr}(\nabla^2_{\text{vec}(\boldsymbol{F}_{l_0})}L ),
\end{align}
where $\bm{V}_p$ denotes the $p$-th column of $\bm{V}$. The gradient of $L$ with respect to $\bm{V}$ is
\begin{align}
    \frac{\partial L}{\partial \bm{V}_p} 
    &= \frac{1}{n} \sum_{i=1}^n \frac{\partial L(f(\boldsymbol{x}_i), \boldsymbol{y}_i)}{\partial f_p(\boldsymbol{x}_i)^T}\frac{\partial f_p(\boldsymbol{x}_i)}{\partial \bm{V}_p} =\frac{1}{n} \sum_{i=1}^n \frac{\partial  L(f(\boldsymbol{x}_i), \boldsymbol{y}_i)}{\partial f_p(\boldsymbol{x}_i)^T}\textup{vec}(\boldsymbol{x}_{L, i})^T,
\end{align}
where $f_p(\boldsymbol{x}_i)$ denotes the $p$-th element in $f(\boldsymbol{x}_i)$.
So, the second-order derivative is as follows:
\begin{align}
    \nabla_{\bm{V}_p}^2L = \frac{1}{n}\sum_{i=1}^n  \frac{\partial^2 L(f(\boldsymbol{x}_i), \boldsymbol{y}_i)}{\partial f_p(\boldsymbol{x}_i)^T \partial f_p(\boldsymbol{x}_i)^T}\textup{vec}(\boldsymbol{x}_{L, i})\textup{vec}(\boldsymbol{x}_{L, i})^T.
\end{align}
The gradient of $L$ with respect to the $p$-th column of $\boldsymbol{W}_{l}$ (denoted by $\boldsymbol{W}_{p, l}$) is :
\begin{align}
    \nabla_{\boldsymbol{W}_{p, l}} L
    = \frac{1}{n} \sum_{i=1}^n \frac{\partial L(f(\boldsymbol{x}_i), \boldsymbol{y}_i)}{\partial f_p(\boldsymbol{x}_i)^T}\left[\bm{V}^T(\prod_{r=l+1}^{L}\Sigma_{r, i} \boldsymbol{W}_{r}^T)\Sigma_{l, i} \boldsymbol{0}_p(\boldsymbol{x}_{l, i})\right]
\end{align}
where $\boldsymbol{0}_p(\boldsymbol{x}_{l, i})$ denotes a $m_{l+1}\times m_{l}$ matrix in which the $p$-th row equals to  the $p$-th row of $\boldsymbol{x}_{l, i}$ and the other  $m_{l+1}-1$ rows equals are all zeros.
And the Hessian of $L$ with respect to $\boldsymbol{W}_{p, l}$ is
\begin{align}
     \nabla_{\boldsymbol{W}_{p, l}}^2 L &=  \frac{1}{n} \sum_{i=1}^n \frac{\partial^2L(f(\boldsymbol{x}_i), \boldsymbol{y}_i)}{\partial f(\boldsymbol{x}_i)^T\partial f(\boldsymbol{x}_i)^T} 
     \left[\bm{V}^T(\prod_{r=l+1}^{L}\Sigma_{r, i} \boldsymbol{W}_{r}^T)\Sigma_{l, i} \boldsymbol{0}_p(\boldsymbol{x}_{l, i})\right]^T \left[\bm{V}^T(\prod_{r=l+1}^{L}\Sigma_{r, i} \boldsymbol{W}_{r}^T)\Sigma_{l, i} \boldsymbol{0}_p(\boldsymbol{x}_{l, i})\right]\\
     & 
     \coloneqq \frac{1}{n} \sum_{i=1}^n \frac{\partial^2L(f(\boldsymbol{x}_i), \boldsymbol{y}_i)}{\partial f(\boldsymbol{x}_i)^T\partial f(\boldsymbol{x}_i)^T} 
       \mathcal{S}^T\bm{V}\bm{V}^T\mathcal{S}
\end{align}
Thus, the trace of Hessian has the same form as \cref{trace1}. So the it is upper bounded by:
\begin{align}
    \text{tr}(\mathbf{H}_{L})
    \leq & \frac{1 }{n}\sum_{i=1}^n \sum_{p=1}^{m_{L+1}} 
    \frac{\partial^2 L(f(\boldsymbol{x}_i), \boldsymbol{y}_i)}{\partial f_p(\boldsymbol{x}_i)^T\partial f_p(\boldsymbol{x}_i)^T }\Vert\textup{vec}(\boldsymbol{x}_{L, i})\Vert^2 + 
    \frac{\Vert\bm{V}\Vert^2}{n}\sum_{l_0=1}^l \sum_{i=1}^n  \text{tr}\left\{ \frac{\partial^2  L(f(\boldsymbol{x}_i), \boldsymbol{y}_i)}{\partial f(\boldsymbol{x}_i)^T \partial f(\boldsymbol{x}_i)^T} 
    \mathcal{S}^T \mathcal{S}\right\},
\end{align}
and lower bounded by:
\begin{align}
    \nabla^2_{\text{vec}(\boldsymbol{F}_l)}L 
    &\geq \frac{1 }{n}\sum_{i=1}^n \sum_{p=1}^{m_{L+1}} 
    \frac{\partial^2 L(f(\boldsymbol{x}_i), \boldsymbol{y}_i)}{\partial f_p(\boldsymbol{x}_i)^T\partial f_p(\boldsymbol{x}_i)^T }\Vert\textup{vec}(\boldsymbol{x}_{L, i})\Vert^2 + \frac{1}{n}\sum_{l_0=1}^l\sum_{i=1}^n \frac{\partial^2  L(f(\boldsymbol{x}_i), \boldsymbol{y}_i)}{\partial f(\boldsymbol{x}_i)^T \partial f(\boldsymbol{x}_i)^T}  \frac{\Vert
     \bm{V}\Vert^2}{\text{tr}(\Lambda^{-1})}
\end{align}

\section{Training Details}
\label{trainingdetails}
The experiment is done in Nvidia 3090 GPU.
For CIFAR-10, the three channels of images are normalized by mean: $(0.4914, 0.4822, 0.4465)$ and standard deviation:$(0.2023, 0.1994, 0.2010)$. 
For CIFAR-100, the three channels of images are normalized by mean: $(0.5071, 0.4865, 0.4409)$ and standard deviation:$ (0.2673, 0.2564, 0.2762)$. 
We use SGD to train different neural networks using batch size 64 for 200 epoches. 
The initial learning rate is set to 0.01 and is divided by 10 at 50, 100, 150 training epoches.
We train using weight decay with 5e-4 and a Nesterov momentum of 0.9 without damping. All images are apply with a simple random horizontal flip.
The Mix-up data augmentation is applied on CIFAR dataset with $\alpha = 1.0$.
For object detection, we following all the default setting of Yolov5 and trained on PASCAL VOC dataset with and without SOS.

\section{Limitation}
\label{limitation section}
SOS depends on the Neural Collapse phenomenon, so currently, it is limited to image tasks. 
For other tasks such as NLP, we leave to future work a more in-depth investigation of these possibilities.
\section{Broader Impacts}
\label{Broader Impacts}
This paper presents work whose goal is to advance the field of Machine Learning. There are many potential societal consequences of our work, none of which we feel must be specifically highlighted here.

\section{More Experiment}
\subsection{Weight Decay Ablation Study}
We also conducted the ablation study of weight decay and shown the result below. 
%
%
From the table below, we can see the similar comparison result of SAM in our main paper: although the accuracy of SOS is lower than weight decay, our SOS can be easily integrated into weight decay and consistently improve the generalization ability. (It is worth noting that in Table 2 in our main paper, both the SGD and SGD + SOS contain weight decay 1e-4. However, in the ablation study below, the SGD denotes SGD without any weight decay.)
\begin{table}[!htb]
    \centering
    \begin{tabular}{l|l|cc|}
    \toprule
        Network                     &  Method         &  Top-1 acc        \\ \midrule
        \multirow{4}{*}{VGG16}      &  SGD &  $68.91_{\pm 0.34}$                \\ 
                                    &  SGD+SOS &  $70.42_{\pm 0.51}$                \\ 
                                    &  SGD+Weight decay &  $72.12_{\pm 0.52}$                \\ 
                                    &  SGD+weight decay + SOS  &  $\mathbf{73.35_{\pm 0.43}}$  \\ \midrule
        \multirow{4}{*}{ResNet18}   &  SGD &  $77.11_{\pm 0.24}$                \\ 
                                    &  SGD+SOS &  $78.31_{\pm 0.33}$              \\ 
                                    &  SGD+Weight decay &  $78.03_{\pm 0.41}$  \\ 
                                    &  SGD+weight decay + SOS  &  $\mathbf{78.43}_{\pm 0.24}$  \\ \midrule
        \multirow{4}{*}{ResNet50}   &  SGD &  $78.18_{\pm 0.15}$                \\ 
                                    &  SGD+SOS &  $79.07_{\pm 0.31}$                \\ 
                                    &  SGD+Weight decay &  $79.30_{\pm 0.22}$                \\ 
                                    &  SGD+weight decay + SOS  &  $\mathbf{79.63}_{\pm 0.27}$  \\ \bottomrule
                                    
    \end{tabular}
    \caption{Apply SOS during training on CIFAR100 dataset.}
    \label{tab:my_label}
\end{table}
\subsection{Trace and Weight Analysis in Training Process}
In this section, we show more result of the weight and Hessian trace analysis during training using cosine learning rate schedule. As Fig.~\ref{sup_w},~\ref{sup_w1},~\ref{sup Hessian during training1},~\ref{sup Hessian during training2} below, we can see that both model weight and Hessian trace with SOS will lower than without SOS.
\newpage
\begin{figure*}[!htb]
    \centering
            \begin{subfigure}{0.31\textwidth}
                \includegraphics[width=1\linewidth]{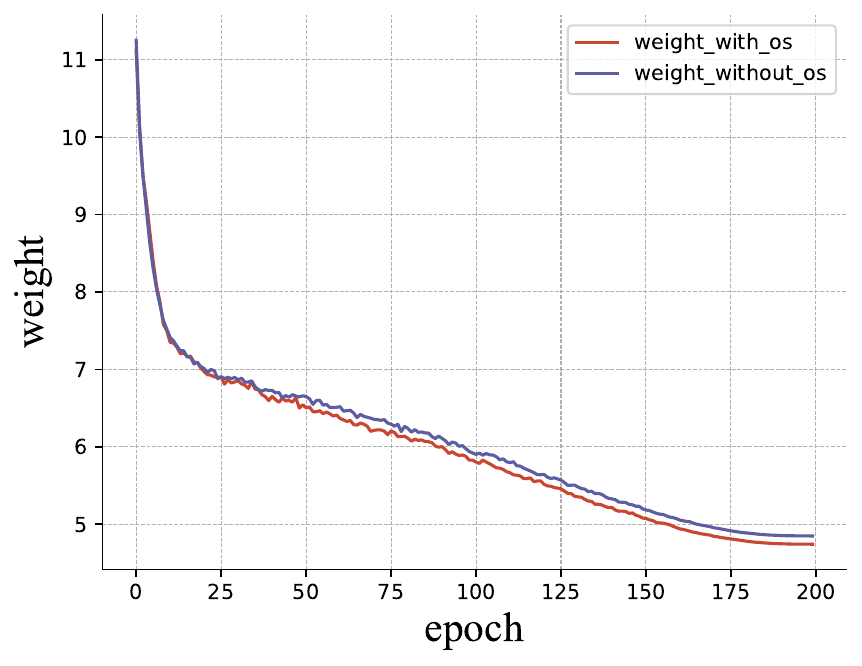}  
                \caption{VGG16 on CIFAR-10}
            \end{subfigure}
            \begin{subfigure}{0.31\textwidth}
                \includegraphics[width=1\linewidth]{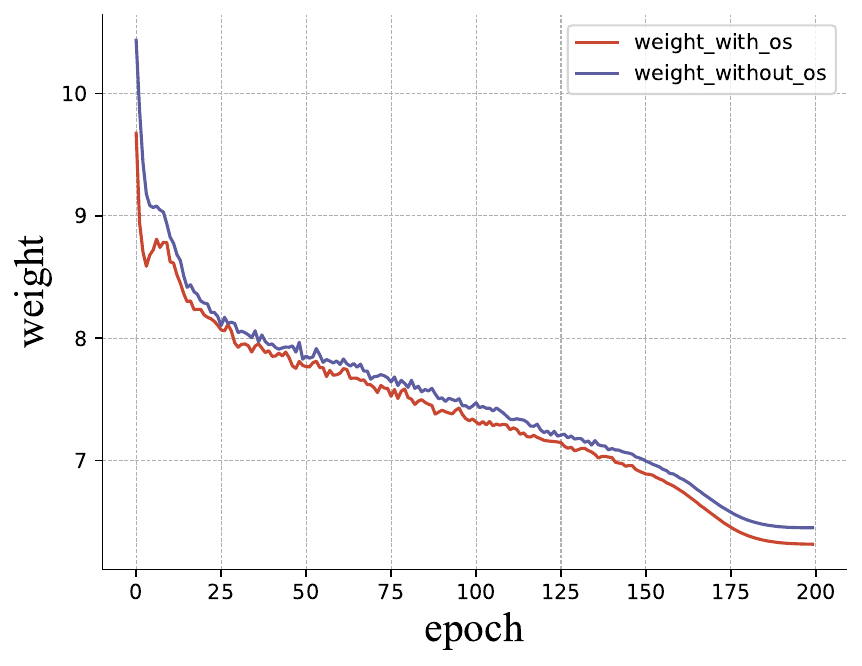}
                \caption{ResNet18 on CIFAR-10}
            \end{subfigure}
            \begin{subfigure}{0.31\textwidth}
                \includegraphics[width=1\linewidth]{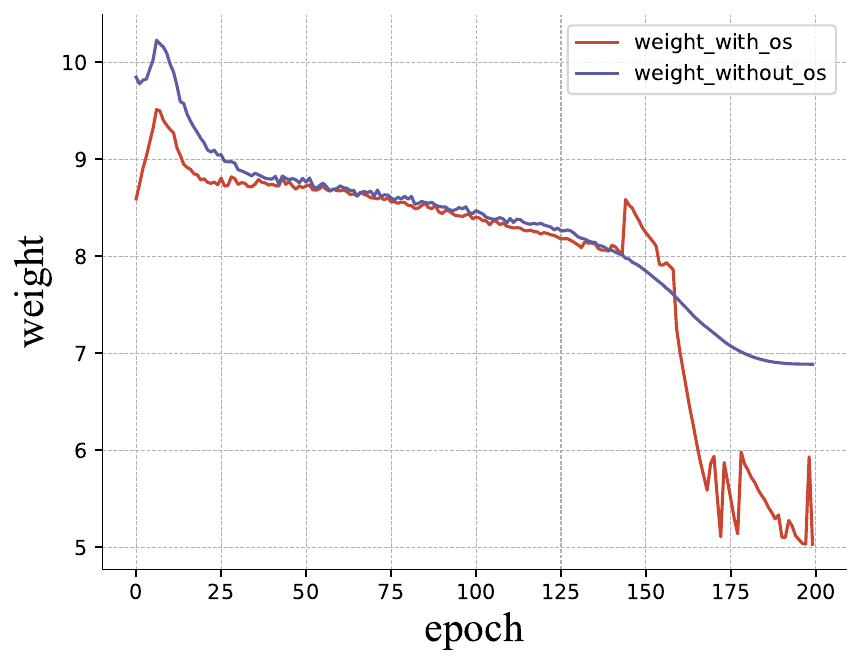}
                \caption{DenseNet121 on CIFAR-10}             
            \end{subfigure}\\
            \begin{subfigure}{0.31\textwidth}
                \includegraphics[width=1\linewidth]{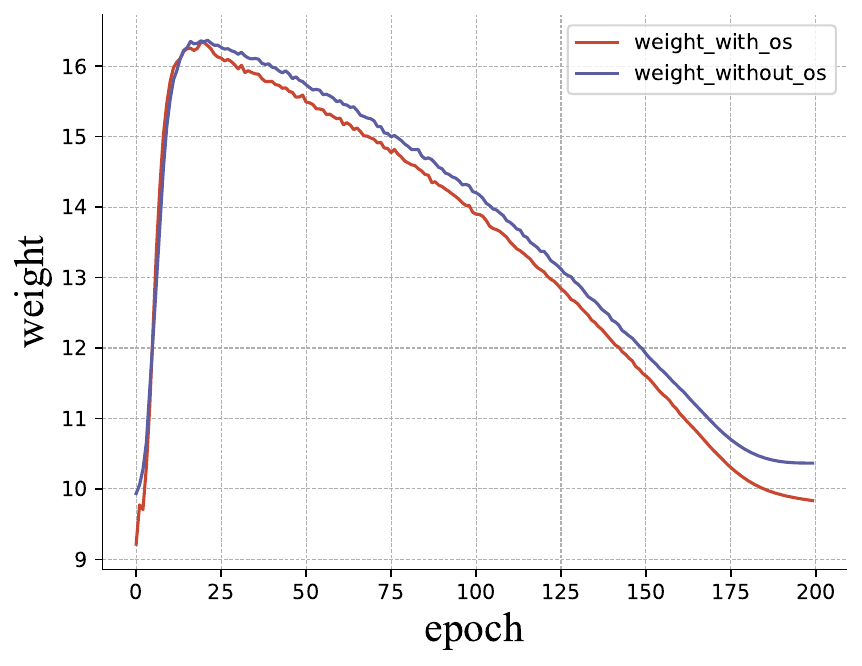}  
                \caption{VGG16 on CIFAR-100}
            \end{subfigure}
            \begin{subfigure}{0.31\textwidth}
                \includegraphics[width=1\linewidth]{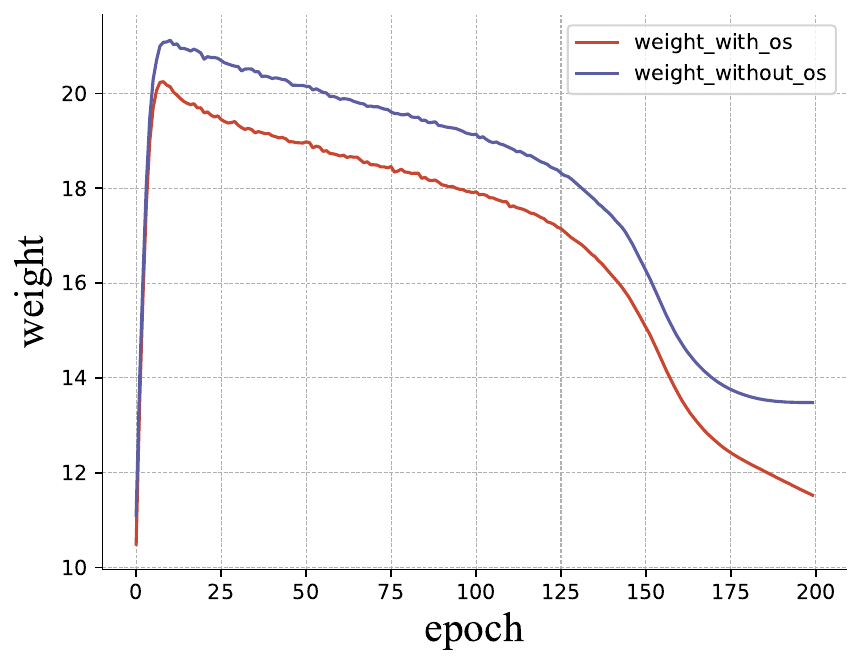}
                \caption{ResNet18 on CIFAR-100}             
            \end{subfigure}
            \begin{subfigure}{0.31\textwidth}
                \includegraphics[width=1\linewidth]{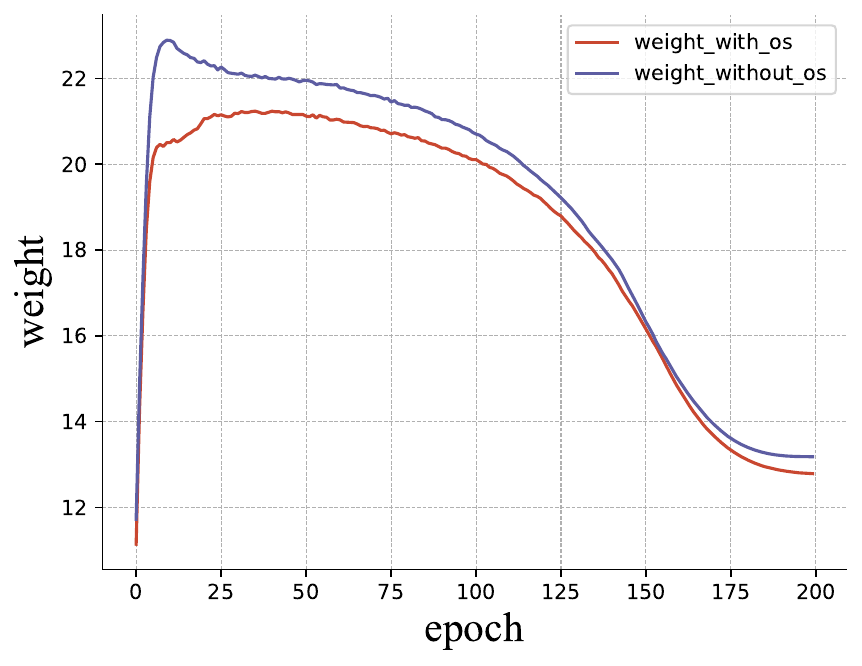}
                \caption{DenseNet121 on CIFAR-100}             
            \end{subfigure}    
            \caption{Weight visualization of different models on CIFAR dataset without mixup during training}
    \label{sup_w}
\end{figure*}
\begin{figure*}[!htb]
    \centering
            \begin{subfigure}{0.31\textwidth}
                \includegraphics[width=1\linewidth]{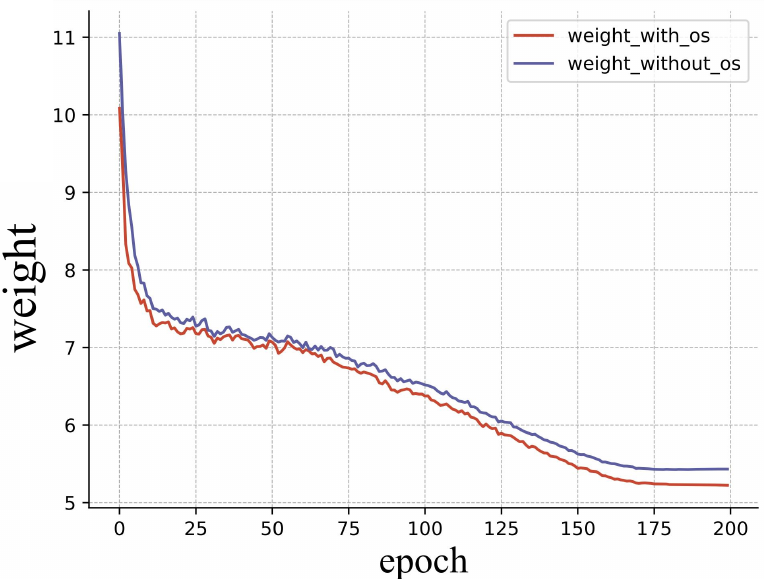}  
                \caption{VGG16 on CIFAR-10}
            \end{subfigure}
            \begin{subfigure}{0.31\textwidth}
                \includegraphics[width=1\linewidth]{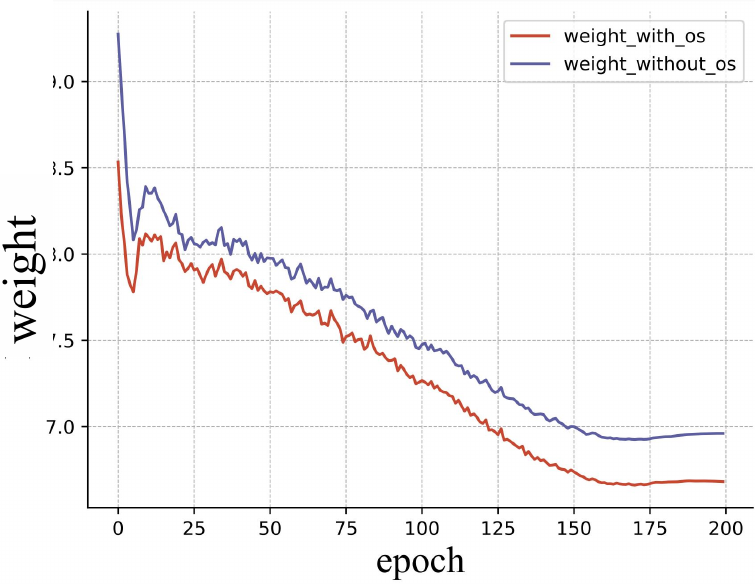}
                \caption{ResNet18 on CIFAR-10}
            \end{subfigure}
            \begin{subfigure}{0.31\textwidth}
                \includegraphics[width=1\linewidth]{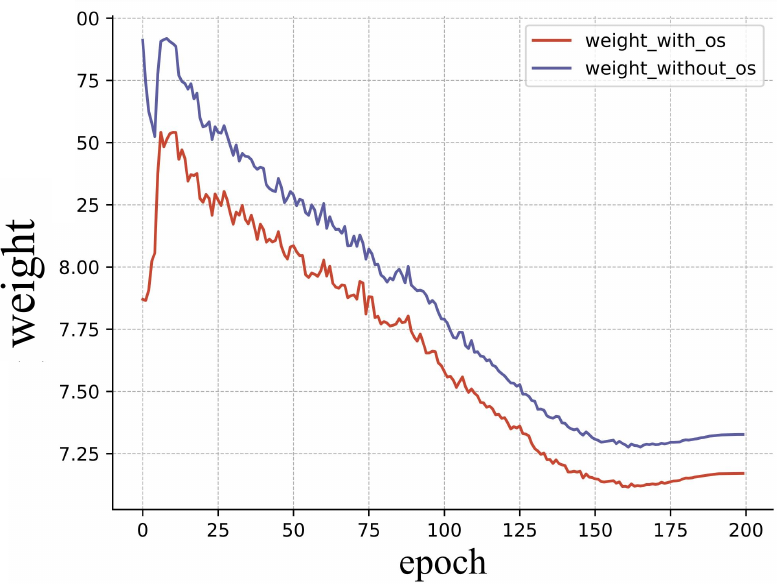}
                \caption{DenseNet121 on CIFAR-10}             
            \end{subfigure}\\
            \begin{subfigure}{0.31\textwidth}
                \includegraphics[width=1\linewidth]{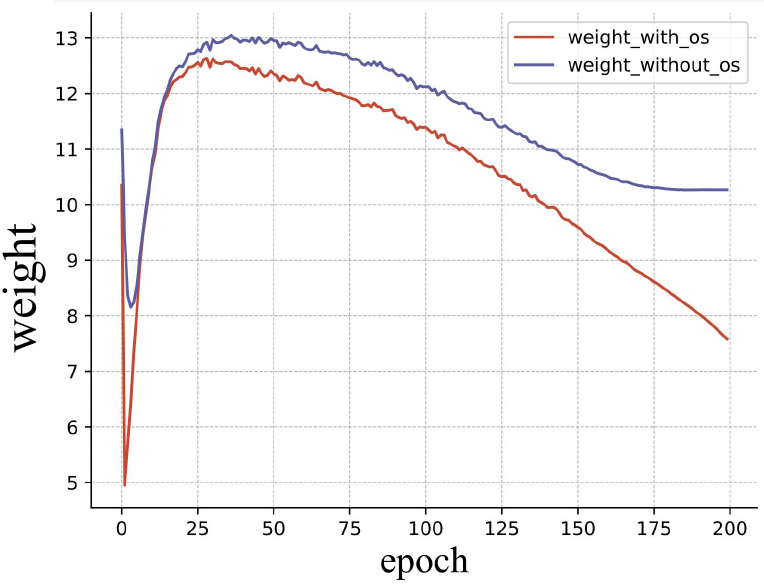}  
                \caption{VGG16 on CIFAR-100}
            \end{subfigure}
            \begin{subfigure}{0.31\textwidth}
                \includegraphics[width=1\linewidth]{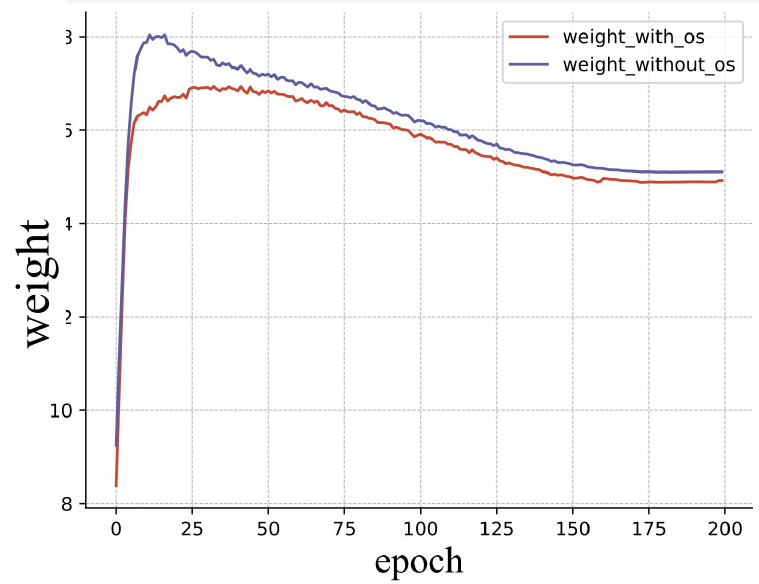}
                \caption{ResNet18 on CIFAR-100}             
            \end{subfigure}
            \begin{subfigure}{0.31\textwidth}
                \includegraphics[width=1\linewidth]{sup_img/crop_d100.pdf}
                \caption{DenseNet121 on CIFAR-100}             
            \end{subfigure}    
            \caption{Weight visualization of different models and dataset with mixup during training}
    \label{sup_w1}
\end{figure*}
\begin{figure*}[!htb]
    \centering
    \begin{subfigure}{0.32\textwidth}
    \centering
    \includegraphics[width=0.95\textwidth]{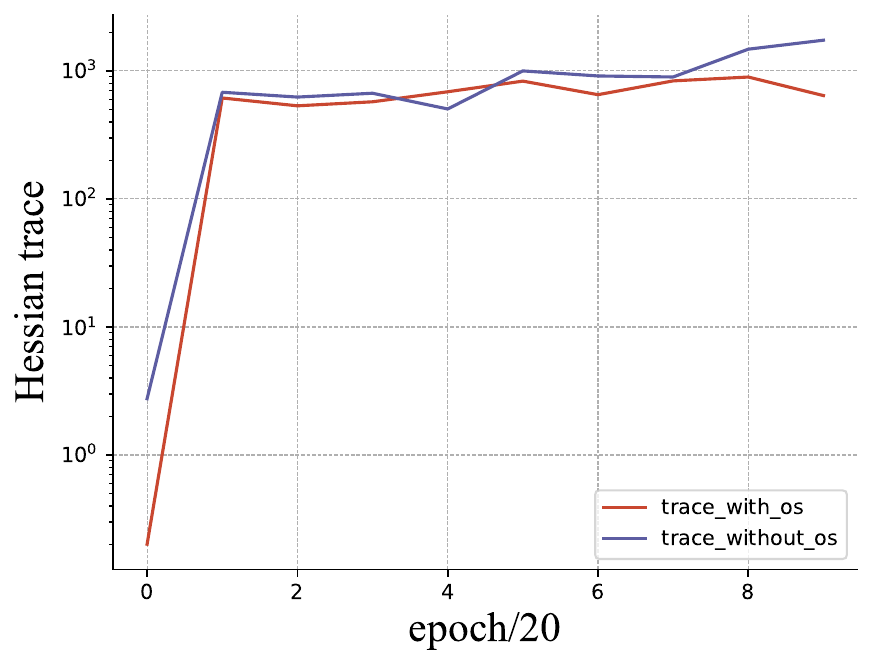}
    \subcaption{VGG16 on CIFAR10}
    \end{subfigure}
    \begin{subfigure}{0.32\textwidth}
    \centering
    \includegraphics[width=0.95\textwidth]{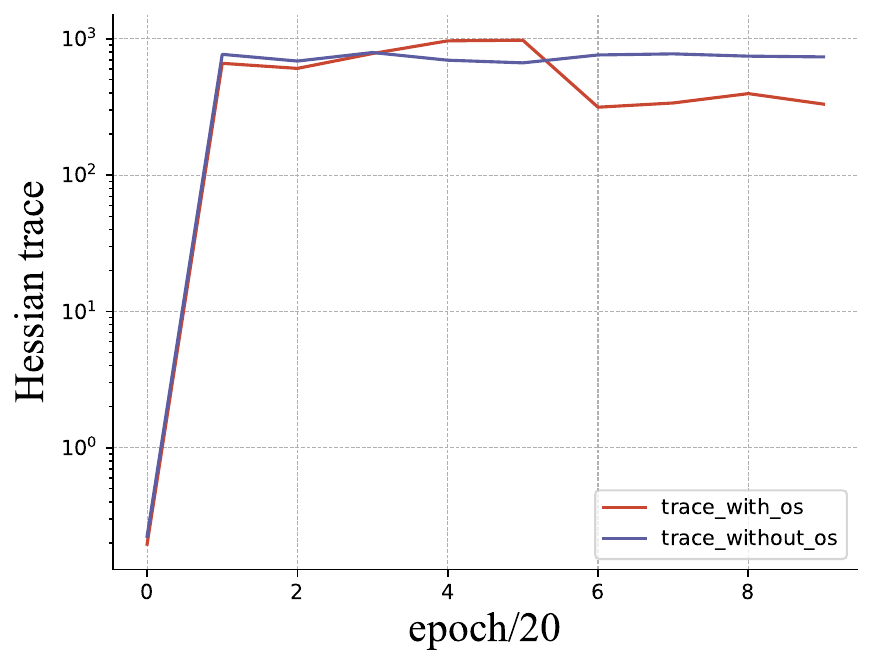}
    \subcaption{ResNet18 on CIFAR10}
    \end{subfigure}
    \begin{subfigure}{0.32\textwidth}
    \centering
    \includegraphics[width=0.95\textwidth]{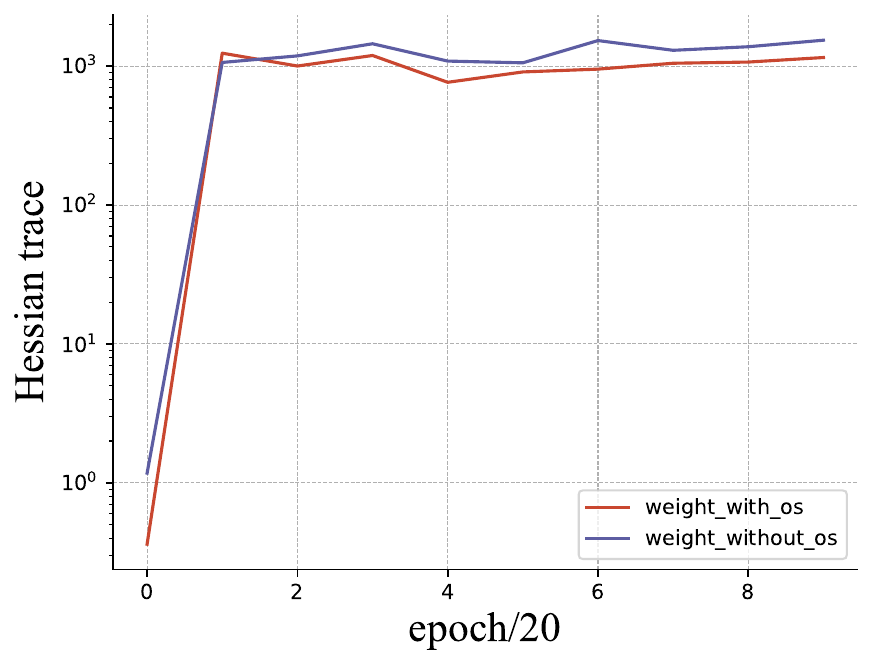}
    \subcaption{DenseNet121 on CIFAR10}
    \end{subfigure}\\
    \begin{subfigure}{0.32\textwidth}
    \centering
    \includegraphics[width=0.95\textwidth]{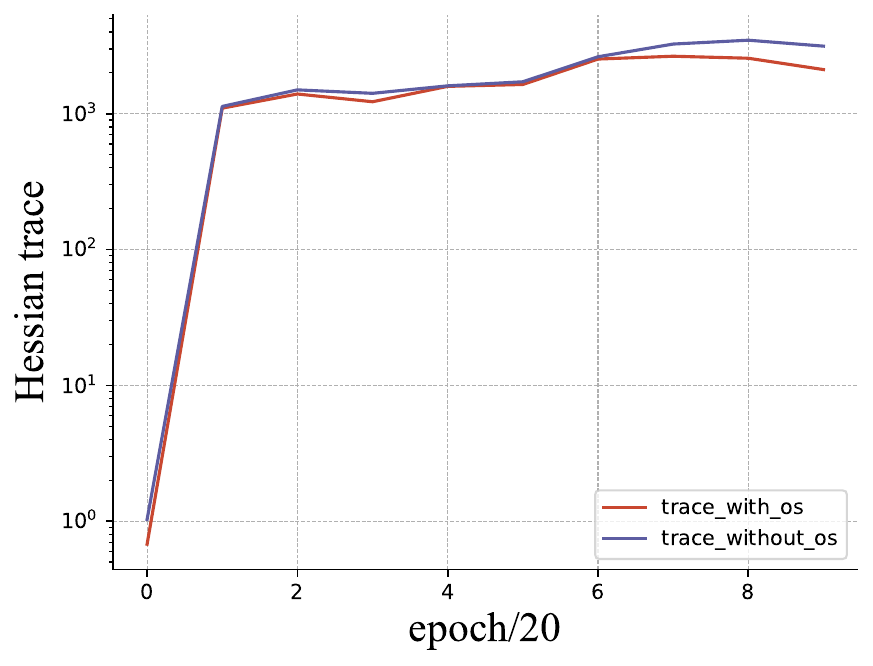}
    \subcaption{VGG16 on CIFAR100}
    \end{subfigure}
    \begin{subfigure}{0.32\textwidth}
    \centering
    \includegraphics[width=0.95\textwidth]{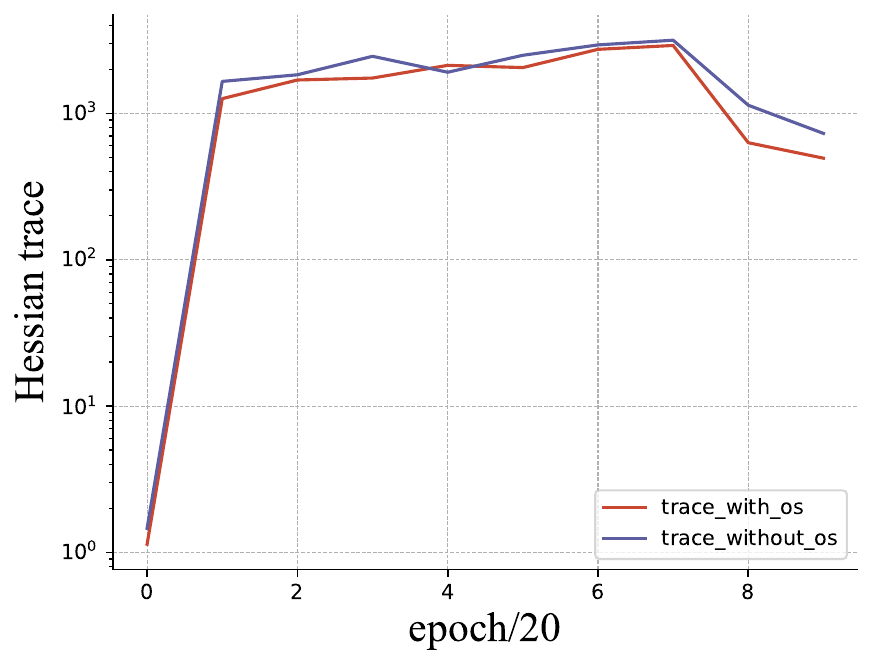}
    \subcaption{ResNet18 on CIFAR100}
    \end{subfigure}
    \begin{subfigure}{0.32\textwidth}
    \centering
    \includegraphics[width=0.95\textwidth]{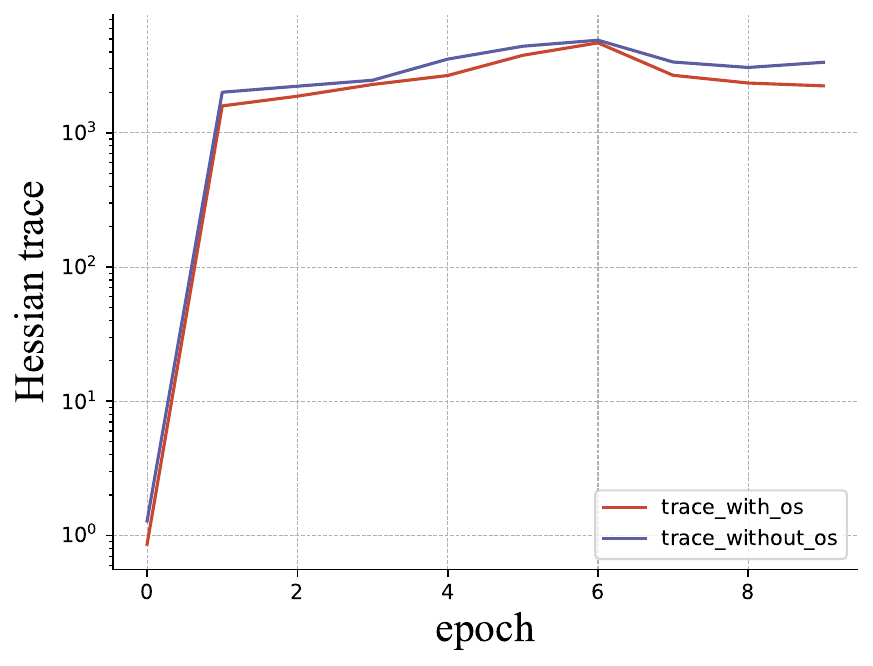}
    \subcaption{DenseNet121 on CIFAR100}
    \end{subfigure}
    \caption{Visualization of the Hessian trace for different models and dataset with mixup during training}
    \label{sup Hessian during training1}
\end{figure*}
\begin{figure*}[!htb]
    \centering
    \begin{subfigure}{0.32\textwidth}
    \centering
    \includegraphics[width=0.95\textwidth]{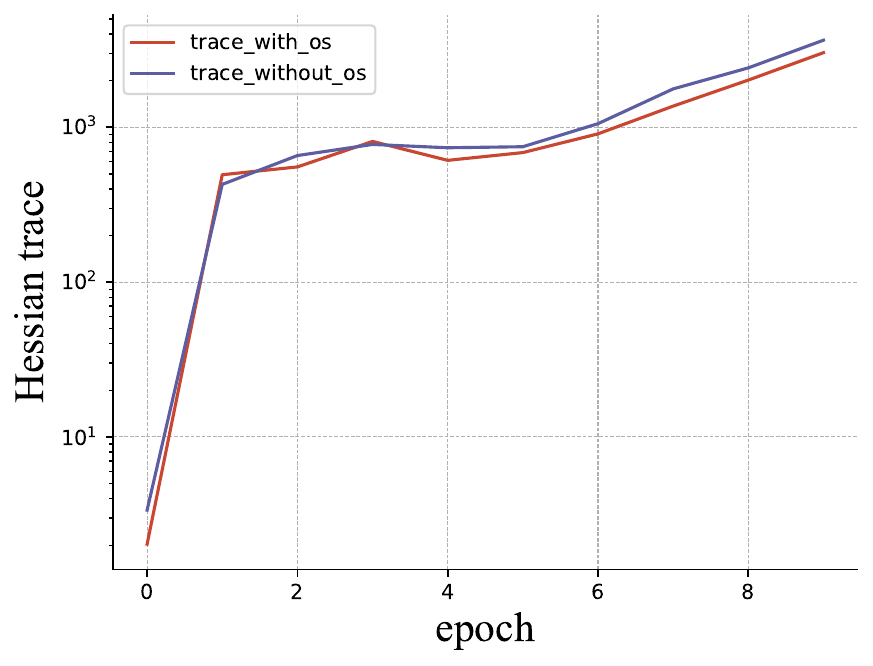}
    \subcaption{VGG16 on CIFAR10}
    \end{subfigure}
    \begin{subfigure}{0.32\textwidth}
    \centering
    \includegraphics[width=0.95\textwidth]{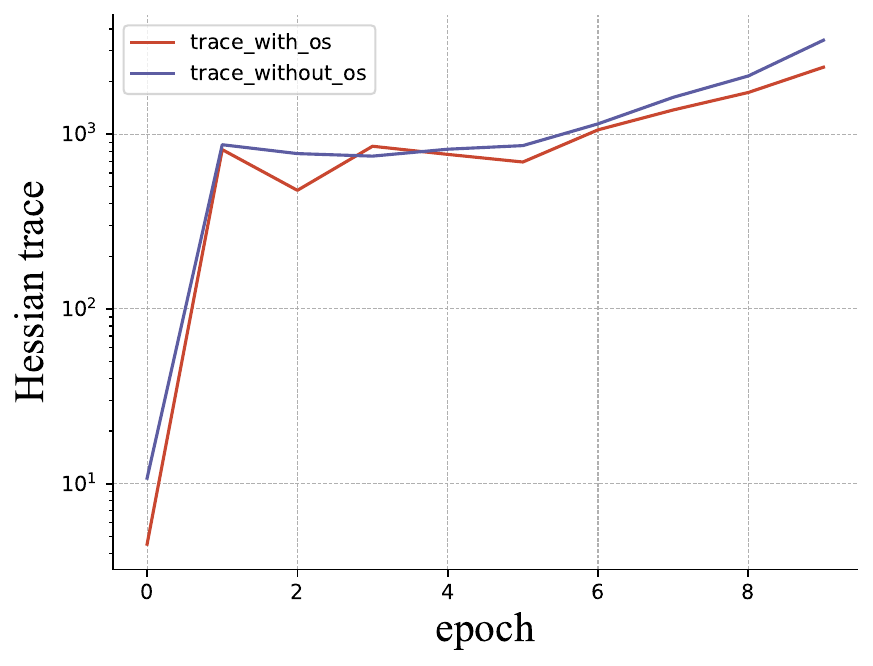}
    \subcaption{ResNet18 on CIFAR10}
    \end{subfigure}
    \begin{subfigure}{0.32\textwidth}
    \centering
    \includegraphics[width=0.95\textwidth]{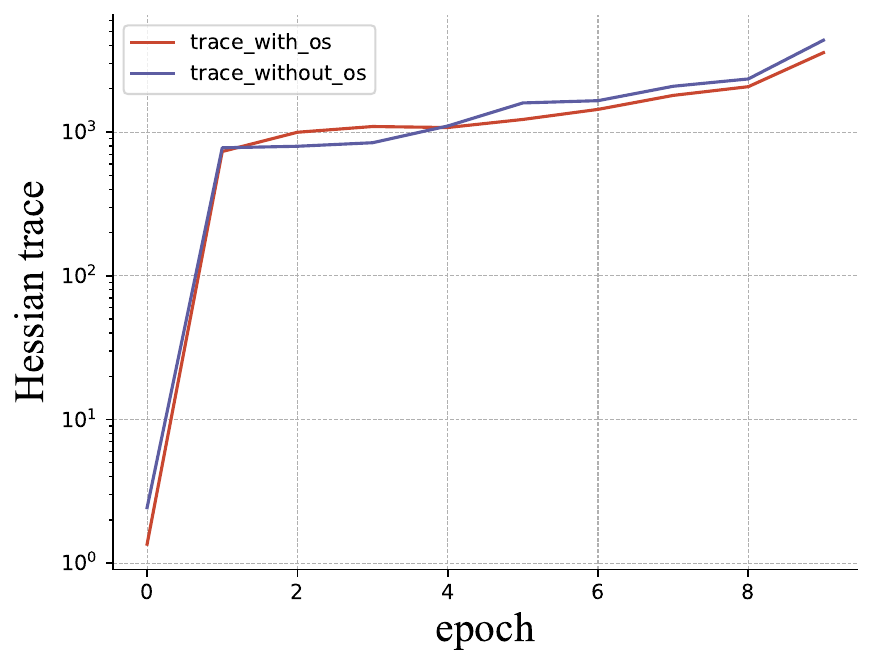}
    \subcaption{DenseNet121 on CIFAR10}
    \end{subfigure}\\
    \begin{subfigure}{0.32\textwidth}
    \centering
    \includegraphics[width=0.95\textwidth]{trace_training/vgg_100_trace.pdf}
    \subcaption{VGG16 on CIFAR100}
    \end{subfigure}
    \begin{subfigure}{0.32\textwidth}
    \centering
    \includegraphics[width=0.95\textwidth]{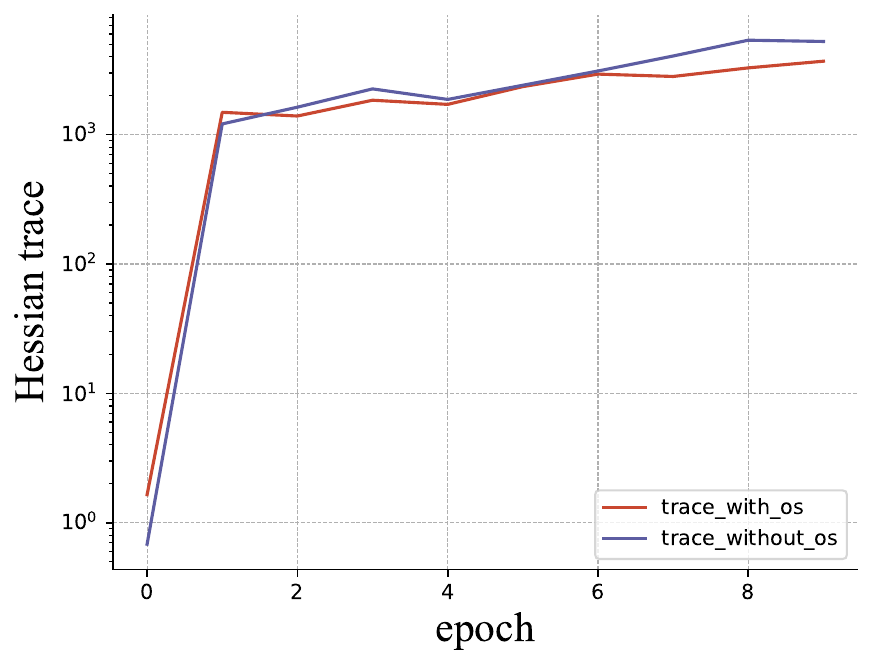}
    \subcaption{ResNet18 on CIFAR100}
    \end{subfigure}
    \begin{subfigure}{0.32\textwidth}
    \centering
    \includegraphics[width=0.95\textwidth]{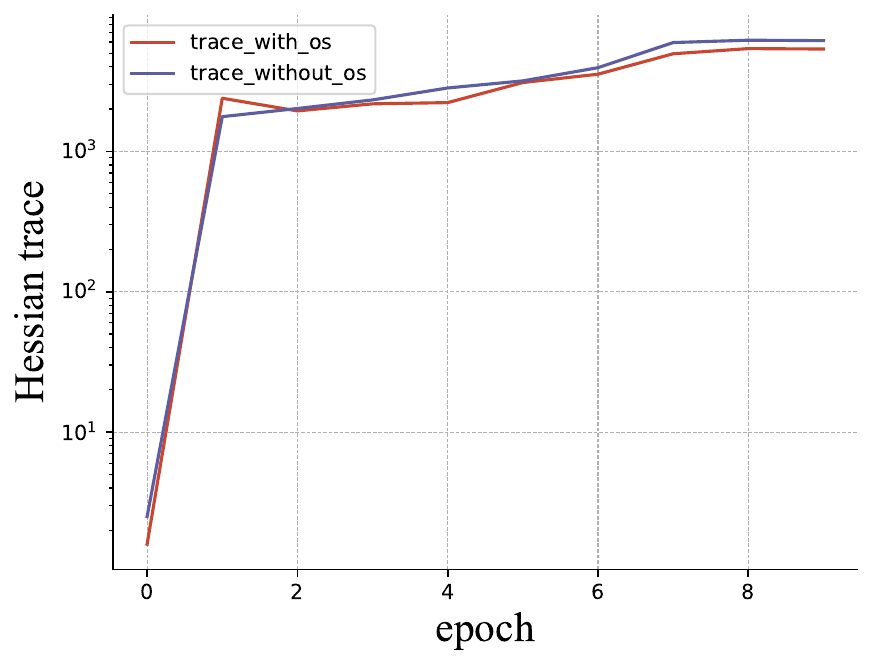}
    \subcaption{DenseNet121 on CIFAR100}
    \end{subfigure}
    \caption{Visualization of the Hessian trace for different models and dataset without mixup during training}
    \label{sup Hessian during training2}
\end{figure*}

\end{document}